\documentclass[letterpaper, 10 pt, conference]{ieeeconf}
\IEEEoverridecommandlockouts
\usepackage{cite}
\usepackage{amsmath,amssymb,amsfonts}
\usepackage{amssymb} 

\usepackage{algorithm}
\usepackage{algpseudocode}
\usepackage{graphicx}
\usepackage{textcomp}
\usepackage{xcolor}
\usepackage{duckuments}
\usepackage{tabularx}
\usepackage{multirow} 
\usepackage{subcaption} 
\usepackage{dblfloatfix}

\def\BibTeX{{\rm B\kern-.05em{\sc i\kern-.025em b}\kern-.08em
    T\kern-.1667em\lower.7ex\hbox{E}\kern-.125emX}}
\begin{document}

\title{Robust Assembly State Reasoning from Action Recognition for Human-Robot Collaboration
}

\author{James Fant-Male$^{1}$ and Roel Pieters$^{1}$
\thanks{$^{1}$Cognitive Robotics group, Unit of Automation Technology and Mechanical Engineering, Tampere University, 33720, Tampere, Finland;
        {\tt\small firstname.surname@tuni.fi}}%
}

\maketitle

\begin{abstract}
 Human Action Recognition (HAR) is frequently investigated in Human-Robot Collaboration (HRC) research to understand what actions have been performed and hence the state of a collaborative task. Accurately tracking an assembly state from HAR is however not fully investigated, and in realistic scenarios is not a trivial task. This research systematically investigates and compares methods for tracking assembly state using action recognition inputs. Investigations using two diverse datasets and five state tracking approaches, including logic-based, Hidden Markov Model (HMM), and neural network (NN) methods, show that optimal approaches are not uniform across different tasks and that different methods fail under different circumstances. Testing is performed using both simulated inputs with varying noise levels and realistic inputs from a HAR model. Results show NN and HMM methods can perform well in tasks with limited variability, but for other scenarios logic-based approaches can be more robust. Methods which model expected action duration are also important for tasks with repeated actions where no additional sensing is provided.
\end{abstract}

\section{Introduction}\label{sec:intro}
Significant developments in Human-Robot Collaboration (HRC) have been made in pursuit of Industry 5.0. The goals of Industry 5.0 highlight the role of human-centred technology~\cite{xu_industry_2021}. In the context of HRC, robot partners should naturally and intuitively understand the user and act accordingly, based on the user's current needs~\cite{fant-male_review_2025}.

A key focus for HRC in manufacturing is the need for the robot partner to know the current assembly stage so it may adaptively plan its future actions and synchronise with the user's current and future requirements. We envisage a common scenario where a collaborative robot arm or mobile platform must decide when to perform certain actions, either consecutively or collaboratively with a user, at the correct time to ensure fluent task completion with minimal wait time. The challenge of understanding which stage of an assembly task the user is performing remains unresolved despite recent research in this area. It is a fundamental aspect of HRC to solve, as without accurate knowledge of the user's current state, efficient future action planning cannot take place. Current approaches tend to either not recognise the current state with reliability and fine granularity in a generalisable manner, or rely on highly specific cues which the robot responds to. These cues reduce the interaction fluidity and add inefficient, cumbersome breaks in the task.

Many developments that work towards assembly stage recognition in the most natural way rely on some form of Human Action Recognition (HAR). Various sensing and computational approaches to HAR have been developed for use in a manufacturing context. However, the integration into HRC scenarios is limited due to a key obstacle. While many research outputs focus on recognition of the action type being performed at a point in time, this is not robustly mapped to the current stage of an assembly. For example, a HAR method may reliably recognise that the operator is screwing in a screw, but in a task containing multiple screwing actions, the exact stage of the assembly is not clear.

Many methods that integrate HAR outputs into a wider HRC manufacturing task tend to rely on primitive logic to decide if a specific action has been completed and to move on to the next. While some investigations incorporate more advanced methods to determine the current state, there lacks a clear comparison of methods and discussion on limitations. Whatever the method is, we propose that it should be both robust and generalisable. For widespread adoption, limiting the use of handcrafted rule-based transitions in favour of a generalisable architecture which can be easily deployed to new assembly tasks should be favoured.

This paper focuses on presenting, investigating and systematically testing varied methods that can be used to determine the state of an assembly task given some current action input. Section~\ref{sec:background} presents the background related literature. Section~\ref{sec:tech_approaches} discusses the problem formulation, datasets used and technical methodologies. Section~\ref{sec:experiments} presents the experiments and results. Section~\ref{sec:discussion} discusses the findings and future work before Section~\ref{sec:conclusion} concludes the paper.

\section{Background} \label{sec:background}
While a significant amount of research has been done on HAR in isolated contexts, both for manufacturing tasks and more widely, limited studies have shown effective integration of HAR in HRC tasks. The robust mapping of human actions to enhance the understanding of assembly state, especially in non-trivial tasks, has yet to be shown effectively. Here we briefly review key works where some form of HAR has been effectively integrated into a collaborative task, where the choice of robot action is based on recognising the user actions and how they are mapped to the current task state.

Demonstration of tasks status recognition to inform action selection and action timing has been shown with a Bayesian network state prediction method. However, the HAR aspect was limited with simple hand proximity to part bins used~\cite{hawkins_probabilistic_2013}. Another study using hand position tracking as action input but with Higher Order Markov Chains for state modelling was used to help a robot decide when to do autonomous actions and when to perform collaborative actions~\cite{zanchettin_prediction_2019}.

An approach using Refined-Multi-Scale Temporal Convolutional Network for HAR based on video data, with 83\% accuracy, and a HMM for assembly intent prediction achieved 90.6\% accuracy when including additional task constraints. The task was however relatively simple, with states directly matched to specific actions~\cite{qu_prediction_2025}. Further work recognising high level actions from atomic actions was performed using an N-gram approach. Using the Assembly-101 dataset, improved accuracy was achieved compared to a HMM method with investigations done using different noise levels to the input~\cite{dwivedi_prediction_2024}. A bi-stream CNN for action recognition was used as input to a variable-length HMM for future action prediction, achieving high accuracy in a machine assembly task~\cite{zhang_hybrid_2021}.

Further work on intent prediction fused action data from an ST-GCN network with facial tracking and object recognition to predict task states and robot actions in a disassembly task~\cite{xiao_intelligent_2025}. HAR using ResNet-34 and LSTM achieved 96.65\% accuracy, which was then combined with a simple state machine to decide robot actions. The experimental validation of the method was however limited~\cite{moutinho_deep_2023}. A novel approach using spiking neural networks has been demonstrated, with HAR, object state and robot state information being encoded as spike signals, with robot action timing predictions from the network. In a simple handover prediction task, the method performed comparably to LSTM and HMM methods~\cite{zhang_fusion-based_2022}.

Previous work using a fusion of IMU and skeleton tracking data for HAR has been shown with an action level LSTM based task prediction method. This was used to predict timing for collaborative actions from the robot on two assembly tasks. The action level LSTM architecture developed will be a key method tested further in this paper~\cite{male_deep_2023}.

One of the most mature demonstrations is the Praxis framework. The method focusses on integrating a HAR framework into a collaborative cell, with efficient methods for recording and labelling data. Different deep-learning based approaches are then used for HAR with demonstration in a collaborative assembly task~\cite{gkournelos_praxis_2024}. A further advanced framework is the SMIRL system, which uses HAR from video and optical flow data to inform state prediction with a state machine system and state transition matrix. A key focus of this work was anomaly detection and action duration recognition for manufacturing insights~\cite{selvaraj_intelligent_2023, selvaraj_real-time_2024}.

There is still a clear gap in identifying robust and generalisable methods to integrate HAR with state prediction and HRC workflows. Many of the tasks in the literature are simple, without repeated actions and minimal deviation from the ideal procedure. Furthermore, many of the approaches require manual handcrafting of state machines and logic for the robot to decide what action to perform next.

\section{Technical Approaches}\label{sec:tech_approaches}
\subsection{Background}
In this work we consider an assembly task, $S$, comprising multiple sequential user actions, $A$. Each action is of primitive type $a$, where multiple of the same primitive type can occur in the task. We assume each task has a linear, non-branching nature, and can thus be simply represented as:

\begin{equation}
    S = \{A_1^a, A_2^a, A_3^a, \ldots, A_n^a\}
\end{equation}

For HRC tasks, accurate task state information is vital for robot action planning. With increasingly conceptual complexity, state information can be thought of in a hierarchy:

\begin{enumerate}
    \item Recognise what type of action, $a$, is being performed, e.g., through HAR.
    \item Recognising which assembly stage, $A_i$, the user is at.
    \item Predicting how long until current/future stages are started/finished.
    \item Finally, which robot actions should be scheduled when for optimal synchronisation.
\end{enumerate}

In this work, we focus on systematically testing key methods for Stage 2, accurately identifying the current stage of a task. We assume some input representing which action type is currently being performed. While we focus on HAR-based inputs, this information may come from any source, e.g. HAR model, vision recognition, tool sensors, etc. In any case, identifying the current assembly stage from this information is non-trivial. The action input is imperfect, as recognition methods such as HAR give inaccurate results, especially in real environments. Also, although the process is ideally a linear, dependent path, this is not the case in real-world situations. Even for simple tasks, users may miss steps, repeat steps or slightly reorder them.

Five methods are investigated here for assembly stage prediction: a baseline logic approach, logic combined with a probabilistic time element, a Hidden Markov Model (HMM), a task-focused Long Short Term Memory (LSTM) network, and multiple action-focused LSTM networks. These methods are presented in more detail in the following subsections.

The approaches investigated have varying implementation requirements. First, they all require a predetermined list of actions which will be progressed through. Secondly, many of the methods assume basic statistics on the action durations, i.e. the mean and standard deviation for each action type in the task, are available. Thirdly, some methods require complete example trial data to learn from. The data requirements are a key consideration in determining the method's usefulness in real life, though in these investigations any data requirements will be met through train and test data splits, as discussed in the following subsection.

\subsection{Dataset Information}\label{sec:datasets}

\begin{table*}[t]
\centering
\caption{Assembly task sequence details. Action sequence ID to name mappings (Note PP = pick and place):
\textbf{HA4M:} 0: N/A, 1: PP Carrier, 2: PP Gear Bearing, 3: PP Planet Gear, 4: PP Carrier Shaft, 5: PP Sun Shaft, 6: PP Sun Gear, 7: PP Sun Gear Bearing, 8: PP Ring Bearing, 9: PP Block 2 on Block 1, 10: PP Cover, 11: PP Screw, 12: PP Allen Key, Turn Screws.
\textbf{IKEA:} 0: N/A, 5: attach shelf to table, 7: flip table, 21: pick up leg, 23: pick up shelf, 31: spin leg.}
\label{tab:dataset_details}
\begin{tabular}{cccccc}
\hline
Dataset & Assembly  & No. Action Primitives & No. States & Action Sequence & No. Train:Val:Test Trials \\
\hline
HA4M & Gear Train   & 13 & 19 & 0, 1, 2, 2, 2, 3, 3, 3, 4, 5, 6, 7, 8, 9, 10, 11, 11, 12, 0 & 151:22:44 \\
IKEA & Coffee Table & 6  & 13 & 0, 21, 31, 21, 31, 21, 31, 21, 31, 7, 23, 5, 0 & 66:10:19 \\
IKEA & Side Table   & 4  & 11 & 0, 21, 31, 21, 31, 21, 31, 21, 31, 7, 0 & 66:10:19 \\
IKEA & TV Bench     & 6  & 13 & 0, 21, 31, 21, 31, 21, 31, 21, 31, 7, 23, 5, 0 & 63:9:19 \\
\hline
\end{tabular}
\end{table*}

For robust testing and analysis across different task types, open-source datasets are used to evaluate the state prediction methods. The datasets must provide labelling of the action primitive at each timestep, have a sequential and determinable assembly sequence, and have skeleton tracking data available which will be used for HAR. Two datasets are identified as being suitable for use and available: the HA4M dataset~\cite{cicirelli2022ha4m} and the IKEA ASM dataset~\cite{ben2021ikea}.

The HA4M dataset consists of 217 trials with 41 participants assembling an epicyclic gear train. The assembly consists of 13 action types, including `null', with actions two and three repeated three times, and action 11 repeated twice, to give 19 states in an assembly sequence when including prior and post null states, as shown in Table~\ref{tab:dataset_details}.

The IKEA ASM dataset contains data on four assembly tasks of different IKEA furniture items. The current assembly status ground truth for this dataset is not so readily available, however can be inferred through processing the action recognition truth output using preprocessing and subsequent manual inspection. A key difference to the HA4M dataset is a larger variation in action sequence between trials, with trials including more natural variation in action order, users correcting errors, repeated actions, etc. Hence, instead of using all actions as part of the task state, only key actions are included in the task state sequence which are required and dependent on each other. Of the 33 original actions recognised, those not corresponding to actions in the state sequence are treated as null. One of the assembly tasks, ``Kallax Shelf Drawer", has an especially large variety in action readings and hence is excluded due to not being able to reliably extract the ground truth assembly status. The other three assembly tasks, ``Lack Coffee Table", ``Lack Side Table" and ``Lack TV Bench", are included with further detail shown in Table~\ref{tab:dataset_details}.

Some of the methods developed require training data and/or data from which to extract action statistics, e.g. mean action durations. The datasets are randomly split into train, validation and test sets by trial, with a 70:10:20 split ratio. The number of trials in each split is shown in Table~\ref{tab:dataset_details}. In all methods presented, any training or input statistics data is taken from the training sets, while all results presented are from the unseen test splits.

A key challenge in the datasets used for this work is that they contain multiple repeated actions. In HA4M, it can be foreseen that the repeated action numbers 2, 3 and 11 present a challenge in distinguishing when one finishes and another starts. Similarly, the IKEA datasets contain multiple repetitions of the same action types in a cyclic manner. This challenge is a key difference from many of the related works which use relatively simple action sequences.

\subsection{Baseline Logic Method}
The baseline method presents a simple, logic-based approach to iterating through the task states. The algorithm, shown in Algorithm \ref{alg:baseline}, is initiated with each action in the sequence set to be neither started nor complete and having time left, $T$, as the mean duration for that action type. At each update step, the current action prediction is compared to that corresponding to the current predicted state. If they match, the time left in the current state is decremented. If the subsequent state action matches the current state action, i.e. the task has a repeated action, and the predicted time left in the current state is 0, then the predicted state is incremented. Additionally, if the previous 3 action inputs match the next state action, the predicted state is incremented.

\begin{algorithm}[tb]
\caption{Baseline State Prediction Update}
\label{alg:baseline}
\begin{algorithmic}[1]
\Require observed action at time $t$, $a^{obs}_t$, predicted state index, $i$, state action $\hat{A}_i$, and remaining action time $T_i$
\State $inc \gets$ False

\If{$\hat{A}_i = A_{i+1}$ \textbf{and} $T_i \le 0$}
    \State $inc \gets$ True
\EndIf

\If{$a^{obs}_t \neq \hat{A}_i$ \textbf{and} {$all(a^{obs}_{t-2}:a^{obs}_t) = A_{i+1}$}}
    \State $inc \gets$ True
    \State $T_{i+1} \gets \max(T_{i+1} - 2\Delta t, 0)$
\EndIf

\If{$inc$}
    \State $\hat{A}_i \gets$ complete
    \State $i \gets i+1$
    \State $\hat{A}_i \gets$ started
\EndIf

\If{$a^{obs}_t = \hat{A}_i$}
    \State $T_i \gets \max(T_i - \Delta t, 0)$
\EndIf

\State \Return $\hat{A}_i$

\end{algorithmic}
\end{algorithm}
\vspace{-2mm}

\subsection{Probabilistic Enhanced Baseline}
The second method enhances the previous baseline with a probabilistic modelling of the current action completion status. As shown in Algorithm \ref{alg:prob_baseline}, similar logic is applied to determine whether to move onto the next state based on the current recognised action. However, the transition to the next state based on the time spent in the current state is determined by drawing from a normal cumulative distribution function (CDF). The normal distribution is modelled with mean, $\mu$, and standard deviation, $\sigma$, coming from the corresponding action durations in the training data. The CDF then provides the probability that the action is complete, given the time spent on the action so far, $T_i$. A threshold of 0.9 is used to trigger moving onto the next action.

\begin{algorithm}[tb] 
\caption{Probabilistic Enhanced Baseline State Update}
\label{alg:prob_baseline}
\begin{algorithmic}[1]
\Require observed action at time $t$, $a^{obs}_t$, predicted state index, $i$, state action $\hat{A}_i$, and time in current state $T_i$

\If{$a^{obs}_t = \hat{A}_i$}
    \State $T_i \gets \max(T_i - \Delta t, 0)$
    \State $\hat{A}_i \gets$ started
\ElsIf{$all(a^{obs}_{t-2}:a^{obs}_t) = A_{i+1}$}
    \State $\hat{A}_i \gets$ complete
    \State $i \gets i+1$
    \State $T_i \gets \max(T_i - 3\Delta t, 0)$
    \State $\hat{A}_i \gets$ started
\ElsIf{$\hat{A}_i = started$ \textbf{and} $\hat{A}_i \neq completed$}
    \State $T_i \gets \max(T_i - \Delta t, 0)$
\EndIf

\If{$\hat{A}_i = started$}
    \State $P(completed) = normal\_cdf(T_i, \mu_{\hat{A}_i}, \sigma_{\hat{A}_i})$
    \If{$P(completed) \geq 0.9$}
        \State $\hat{A}_i \gets$ complete
        \State $i \gets i+1$
        \State $T_i \gets 0$
        \State $\hat{A}_i \gets$ started
    \EndIf
\EndIf

\State \Return $\hat{A}_i$

\end{algorithmic}
\end{algorithm}

\subsection{Hidden Markov Model}
Hidden Markov Model (HMM) based approaches are popular for modelling task states and are seen in various collaborative assembly examples~\cite{zanchettin_prediction_2019, zhang_hybrid_2021}. HMMs are well-suited for estimating state transitions from indirect observations. The hmmlearn Python library is used to implement a CategoricalHMM with predefined transition and emission matrices to avoid the need to train from data. Given the known sequential task structure and action types, this knowledge can be incorporated directly into the model without requiring training data. The number of hidden states, $N$, is modelled as the number of actions in the sequence, and the number of features, $K$, as the number of possible action readings. The starting probability vector $\boldsymbol{\pi}$ is initialised so that the model always begins in the first state:

\[
\boldsymbol{\pi} =
\begin{bmatrix}
1 & 0 & 0 & \cdots & 0
\end{bmatrix}
\]

The transition matrix, $A \in \mathbb{R}^{N \times N}$, is defined such that the probability of remaining in the same state is $0.7$, the next $0.3\times0.75=0.225$ and the next $0.3\times0.25=0.075$, with the final state being absorbing:

\[
A =
\begin{bmatrix}
0.7 & 0.225 & 0.075 & 0 & \cdots & 0 \\
0   & 0.7   & 0.225 & 0.075 & \cdots & 0 \\
0   & 0     & 0.7   & 0.0225 & \cdots & 0 \\
\vdots & \vdots & \vdots & \vdots & \ddots & 0.3 \\
0 & 0 & 0 & 0 & \cdots & 1
\end{bmatrix}
\]

The emission matrix $B \in \mathbb{R}^{N \times K}$ models the probability of observing each action type from each state. Each state emits its corresponding observation with probability $0.7$, and the remaining probability is distributed uniformly among the other observation types:
\[
B_{i,j} =
\begin{cases}
0.7 & \text{if observation } j \text{ corresponds to state } i \\
\frac{1-0.7}{K-1} & \text{otherwise}
\end{cases}
\]



\subsection{LSTM Task Model}
Neural Network based learning approaches are well known for being able to learn complex relationships between data, with Long Short-Term Memory (LSTM) networks enabling learning and prediction over time series data. Two different approaches using LSTM networks are presented in this work: task-based modelling and action-based modelling.

For task-based modelling, an LSTM network is trained to learn the characteristics of a specific task and output the current state. Specifically, a network is built consisting of 2 layers of 50 LSTM units followed by a fully connected layer with 50 units and ReLU output, before final output with shape equal to the number of states in the sequence. The input shape is the number of possible action types. The inputs and outputs are therefore the one-hot encoded vectors corresponding to the action prediction from HAR (input) and action sequence index (output).

For each task type, a model is trained using data for that task from the training set. The input is taken as the sequence of actions corresponding to the ideal output state sequence, with random noise applied at a rate of 20\% to simulate imperfect sensor readings. The network is trained on sequences of 5\,s of data, with 1\,s step between data windows. Training is performed using the Adam optimiser with cross-entropy loss. Learning rate is set to 0.001 for 15 epochs and batch size of 128. 

\begin{figure*}[b]
    \centering
    \begin{subfigure}{0.495\linewidth}
        \centering
        \includegraphics[width=\linewidth]{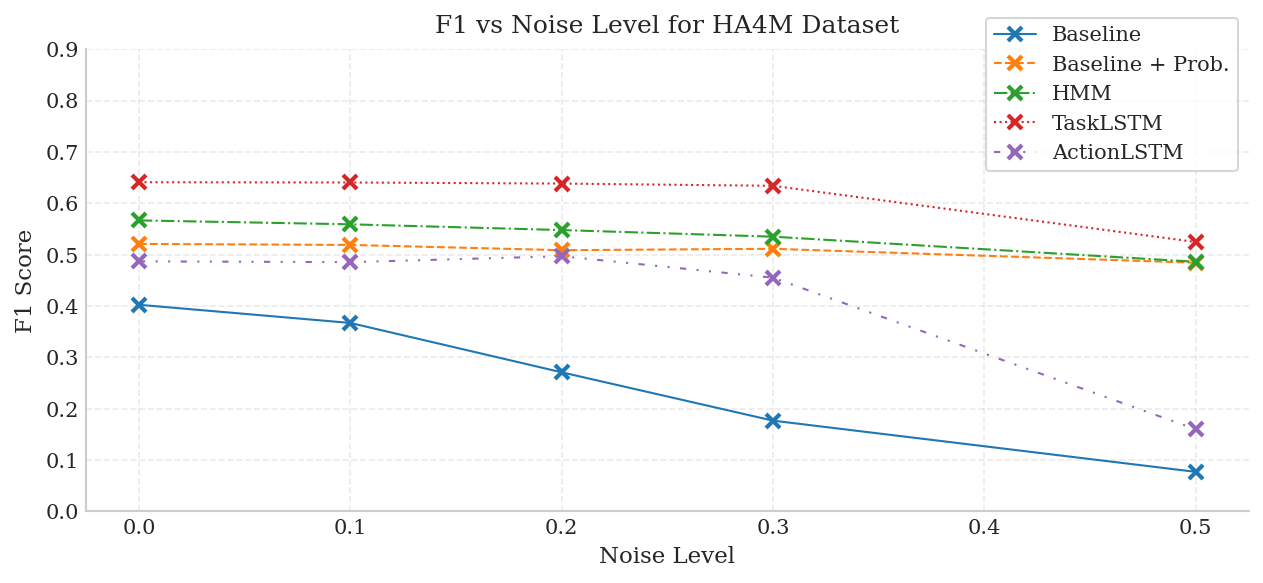}
        \label{fig:f1_noise_ha4m}
    \end{subfigure}
    \hfill
    \begin{subfigure}{0.495\linewidth}
        \centering
        \includegraphics[width=\linewidth]{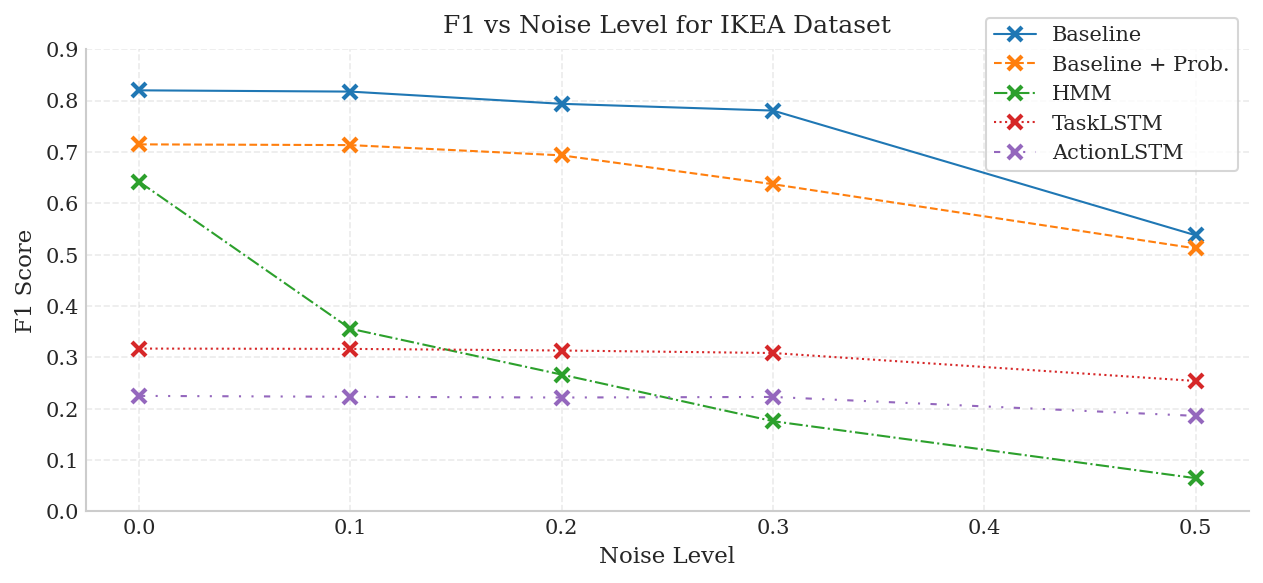}
        \label{fig:f1_noise_ikea}
    \end{subfigure}
    \vspace{-5mm}
    \caption{State prediction F1 score with different noise level applied to action input for (a) HA4M and (b) IKEA datasets.}
    \label{fig:f1_with_noise}
\end{figure*}
\subsection{LSTM Action Model}
\begin{figure}[t]
    \centering
    \includegraphics[trim={1mm 1mm 1mm 1mm},clip, width=\linewidth]{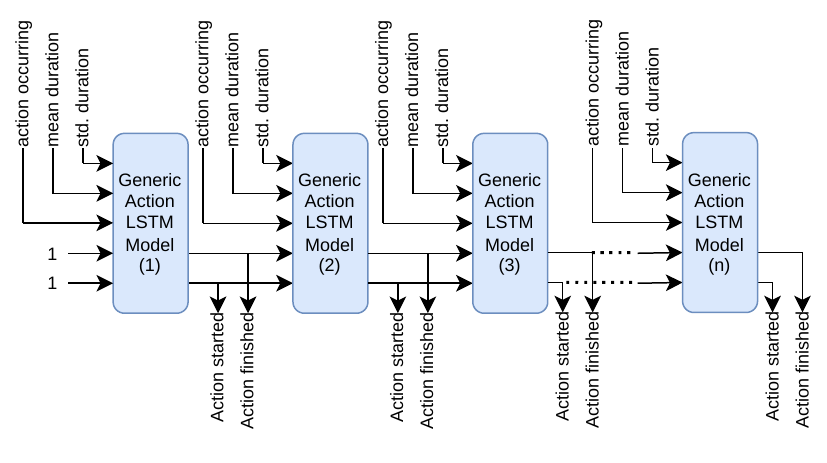}
    \caption{Action agnostic LSTM network architecture with adjustment based on action specific input parameters.}
    \vspace{-3mm}
    \label{fig:lstm_chain_structure}
\end{figure}

The second LSTM based approach focuses on action-based modelling. As previously presented in~\cite{male_deep_2023}, the architecture aims to avoid training a model for each new assembly task which requires significant data and time. Instead, a single, generic action predictor network is trained which can then be instantiated and chained together for each action in a task. The model is trained to output the probability that the specific action for which it is instantiated has both started and been completed. The model takes as input whether the action type for that instantiation is currently being performed, the mean and standard deviation duration for the action type, and the outputs from the previous model in the sequence, i.e. if the previous action has started and been completed. This network structure is depicted in Fig.~\ref{fig:lstm_chain_structure}. 

The action predictor model is structured with 2 layers of 50 LSTM units, followed by a fully connected layer of 50 units and ReLU activation. The model is trained using data from all task types and action types across the two datasets to create a generic model capable of making predictions for different types of action. The model is trained for 2 epochs using the Adam optimiser, binary cross entropy loss, learning rate of 0.0001 and batch size of 128. As used previously, 5\,s data sequences are used with 20\% noise added to the ideal action input sequence.

Each action started and completed prediction for the task sequence is converted to binary by thresholding at 0.5. The current action state is taken as the last action marked as started but not completed. If none match, it is taken as the action after the last one marked as started and completed.

\section{Experiments}\label{sec:experiments}
\subsection{Simulated Data Experiments}\label{sec:simulated_data}

A two-stage experimentation approach is followed. Preliminary testing systematically investigates the performance of the methods under varying input quality. This is achieved by taking the action corresponding to the final desired output state and using this as the input action to the model at each time step. Increasing levels of noise are applied to this by randomly changing the input action type to another action type to simulate a less than perfect HAR system. Testing is done on both the HA4M dataset and three tasks from the IKEA ASM dataset presented in Section~\ref{sec:datasets}. Average results across the test trials can be seen in Table~\ref{tab:results_noisy}, with plots of changing F1 score with noise level in Fig.~\ref{fig:f1_with_noise}.

It can be seen that different methods perform best for different datasets. For the HA4M dataset, the TaskLSTM performs best with F1 of 0.64 for noise levels of 0-0.3, and the HMM is the best method which does not require training. For the IKEA dataset, the two Baseline methods perform highest, while the HMM achieves 0.64 F1 at 0\% noise before falling away quickly as noise is added.

The confusion matrices in Figs.~\ref{fig:conf_mats_ha4m_nl0.0}-\ref{fig:conf_mats_Lack_Side_Table_nl0.3} give further insight into why the methods are failing. Values in the lower left of the matrices indicate the model falling behind the task state. It can be seen the Baseline predictor is prone to getting stuck in a state and not moving on. The HMM struggles to distinguish between sequential actions of the same type in the HA4M dataset, given the lack of action duration modelling. In the IKEA task, the HMM is prone to jumping ahead to the end of the task, while the actionLSTM becomes stuck after the fourth action.

\begin{table}[tb]
    \centering
    \caption{State recognition metrics with different input noise levels for test trials. Data shown for HA4M task, average of IKEA tasks and average across all tasks.}
    \resizebox{\linewidth}{!}{%
    \begin{tabular}{cc|cc|cc|cc}
         & & \multicolumn{6}{c}{Task Type} \\
        \multirow{2}{*}{\shortstack[l]{Noise\\Level}} &
        & \multicolumn{2}{c|}{HA4M} 
        & \multicolumn{2}{c|}{IKEA ASM} 
         & \multicolumn{2}{c}{Average} \\
         & Method 
        & Acc & F1 
        & Acc & F1 
        & Acc & F1 \\ \hline

        \multirow{5}{*}{0.0}
        & Baseline & 41.3 & 40.2 & \textbf{83.3} & \textbf{82.0} & \textbf{72.8} & \textbf{71.6} \\
        & Baseline + Prob & 51.4 & 52.1 & 71.7 & 71.5 & 66.6 & 66.6 \\
        & HMM & 59.2 & 56.7 & 58.4 & 64.2 & 58.6 & 62.3 \\
        & Task LSTM & \textbf{63.1} & \textbf{64.1} & 40.4 & 31.7 & 46.1 & 39.8 \\
        & Action LSTM & 48.7 & 48.7 & 21.9 & 22.5 & 28.6 & 29.0 \\ \hline

        \multirow{5}{*}{0.1}
        & Baseline & 38.1 & 36.7 & \textbf{83.0} & \textbf{81.8} & \textbf{71.8} & \textbf{70.5} \\
        & Baseline + Prob & 51.2 & 51.9 & 71.5 & 71.4 & 66.4 & 66.5 \\
        & HMM & 58.4 & 55.9 & 35.8 & 35.6 & 41.5 & 40.7 \\
        & Task LSTM & \textbf{63.1} & \textbf{64.1} & 40.2 & 31.7 & 45.9 & 39.8 \\
        & Action LSTM & 48.6 & 48.5 & 21.6 & 22.3 & 28.4 & 28.9 \\ \hline

        \multirow{5}{*}{0.2}
        & Baseline & 28.2 & 27.1 & \textbf{80.8} & \textbf{79.4} & \textbf{67.6} & \textbf{66.3} \\
        & Baseline + Prob & 50.1 & 50.9 & 69.5 & 69.4 & 64.7 & 64.7 \\
        & HMM & 57.2 & 54.8 & 27.9 & 26.6 & 35.2 & 33.7 \\
        & Task LSTM & \textbf{62.9} & \textbf{63.9} & 39.7 & 31.3 & 45.5 & 39.5 \\
        & Action LSTM & 49.3 & 49.7 & 21.3 & 22.2 & 28.3 & 29.1 \\ \hline

        \multirow{5}{*}{0.3}
        & Baseline & 19.1 & 17.7 & \textbf{79.8} & \textbf{78.1} & \textbf{64.6} & \textbf{63.0} \\
        & Baseline + Prob & 50.4 & 51.2 & 64.1 & 63.7 & 60.7 & 60.6 \\
        & HMM & 55.7 & 53.5 & 19.5 & 17.6 & 28.6 & 26.6 \\
        & Task LSTM & \textbf{62.4} & \textbf{63.4} & 39.1 & 30.8 & 45.0 & 39.0 \\
        & Action LSTM & 44.6 & 45.5 & 21.4 & 22.3 & 27.2 & 28.1 \\ \hline

        \multirow{5}{*}{0.5}
        & Baseline & 9.0 & 7.7 & \textbf{57.2} & \textbf{53.8} & 45.2 & 42.3 \\
        & Baseline + Prob & 47.3 & 48.4 & 51.8 & 51.2 & \textbf{50.7} & \textbf{50.5} \\
        & HMM & \textbf{50.0} & 48.6 & 7.8 & 6.5 & 18.4 & 17.0 \\
        & Task LSTM & 49.0 & \textbf{52.5} & 33.5 & 25.4 & 37.4 & 32.2 \\
        & Action LSTM & 14.6 & 16.0 & 17.5 & 18.6 & 16.7 & 17.9 \\

    \end{tabular}%
     }
     \vspace{-3mm}
    \label{tab:results_noisy}
\end{table}

\newcommand{\confmatsize}{0.19}
\begin{figure*}
    \centering
    \begin{subfigure}{\confmatsize\linewidth}
        \centering
        \includegraphics[width=\linewidth]{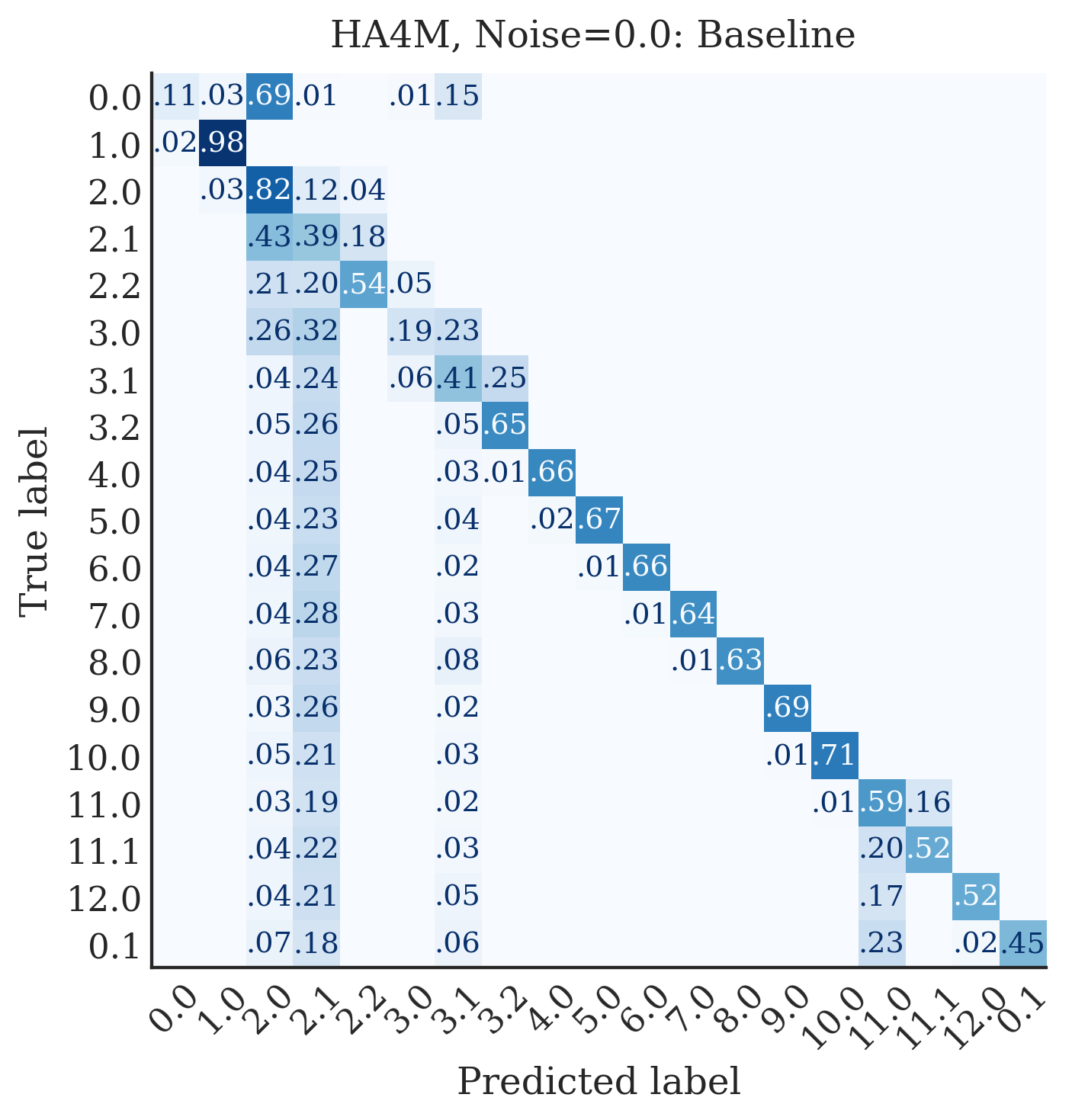}
    \end{subfigure}
    \hfill
    \begin{subfigure}{\confmatsize\linewidth}
        \centering
        \includegraphics[width=\linewidth]{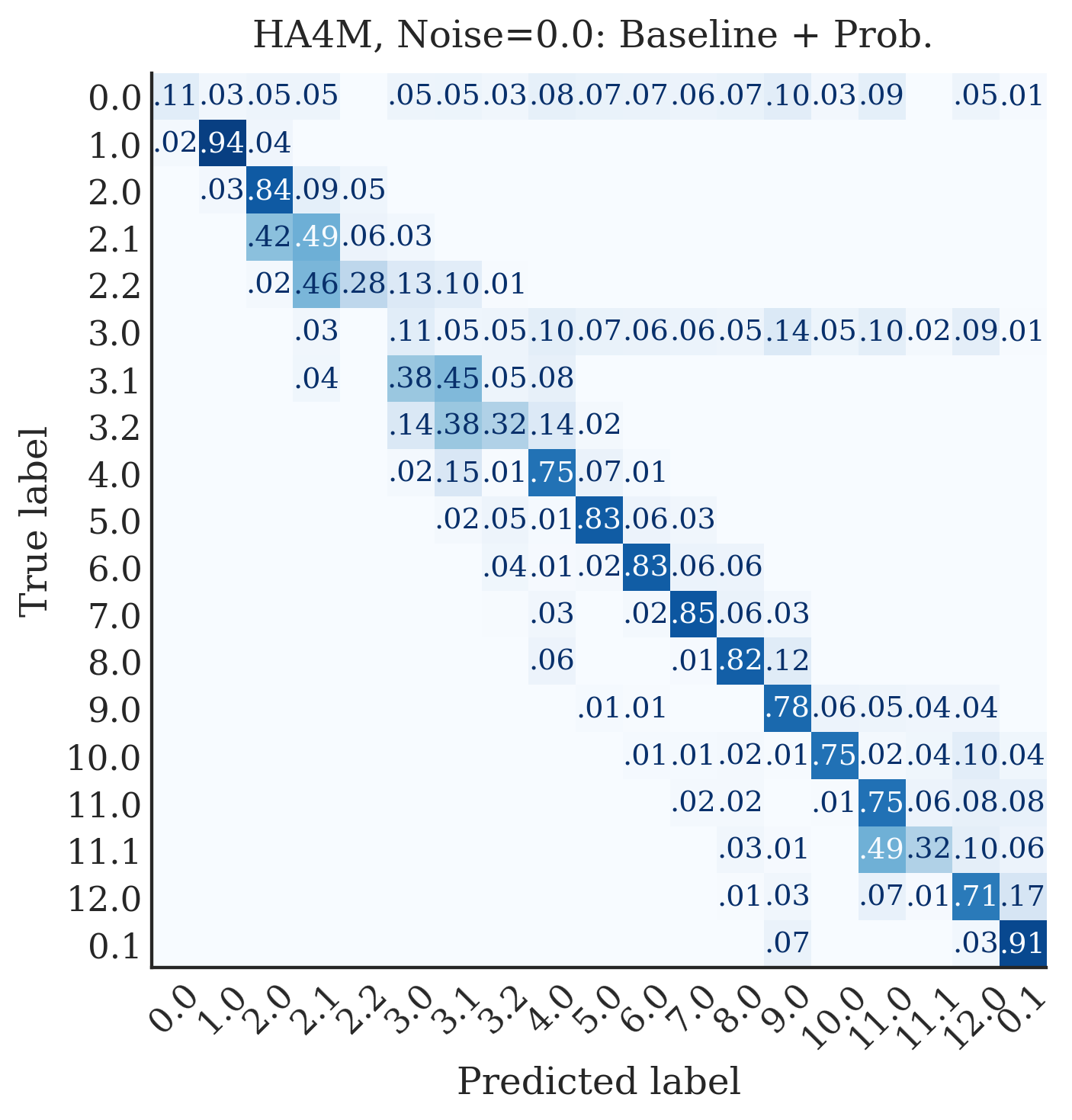}
    \end{subfigure}
    \hfill
    \begin{subfigure}{\confmatsize\linewidth}
        \centering
        \includegraphics[width=\linewidth]{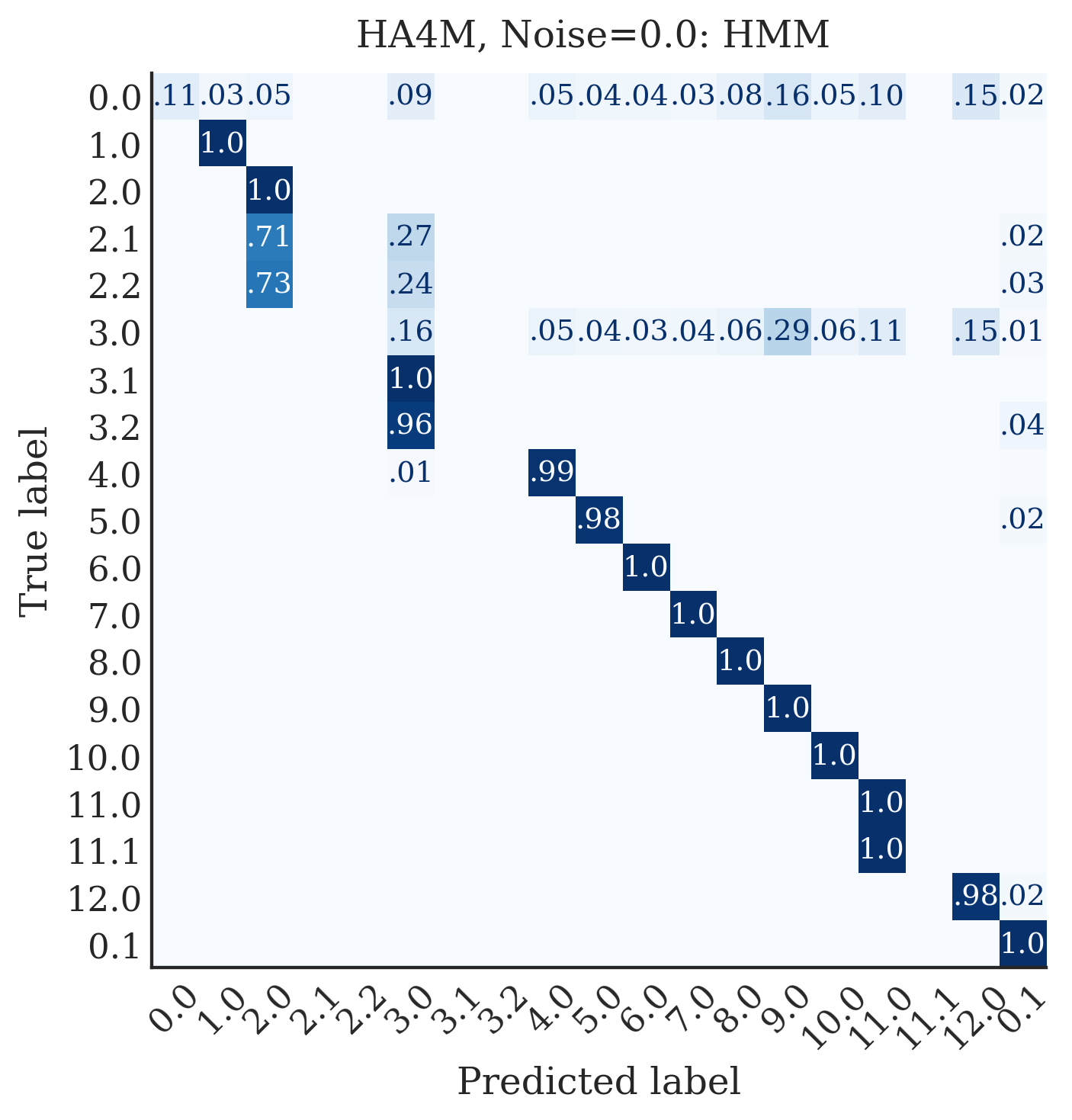}
    \end{subfigure}
    \hfill
    \begin{subfigure}{\confmatsize\linewidth}
        \centering
        \includegraphics[width=\linewidth]{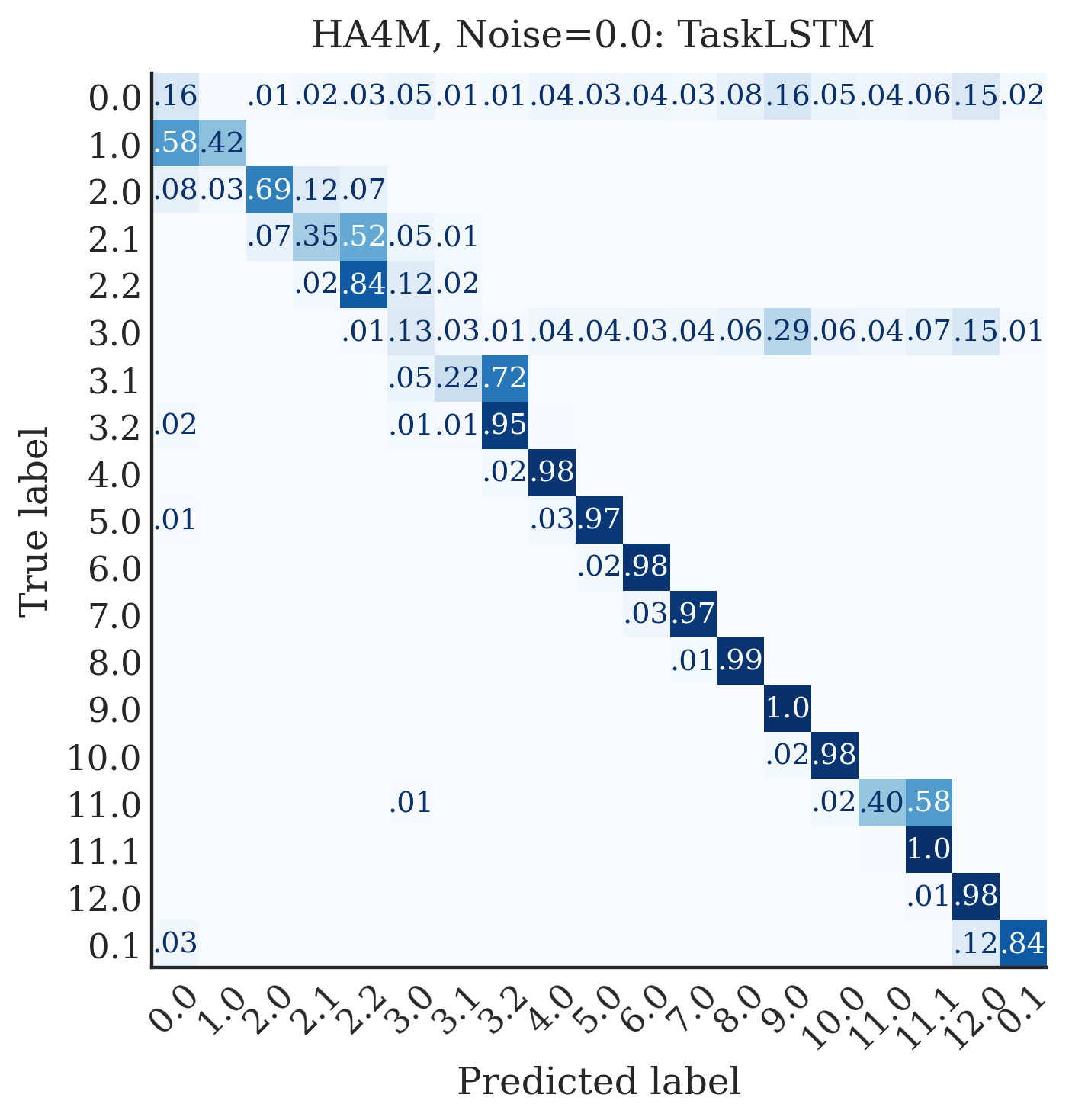}
    \end{subfigure}
    \hfill
    \begin{subfigure}{\confmatsize\linewidth}
        \centering
        \includegraphics[width=\linewidth]{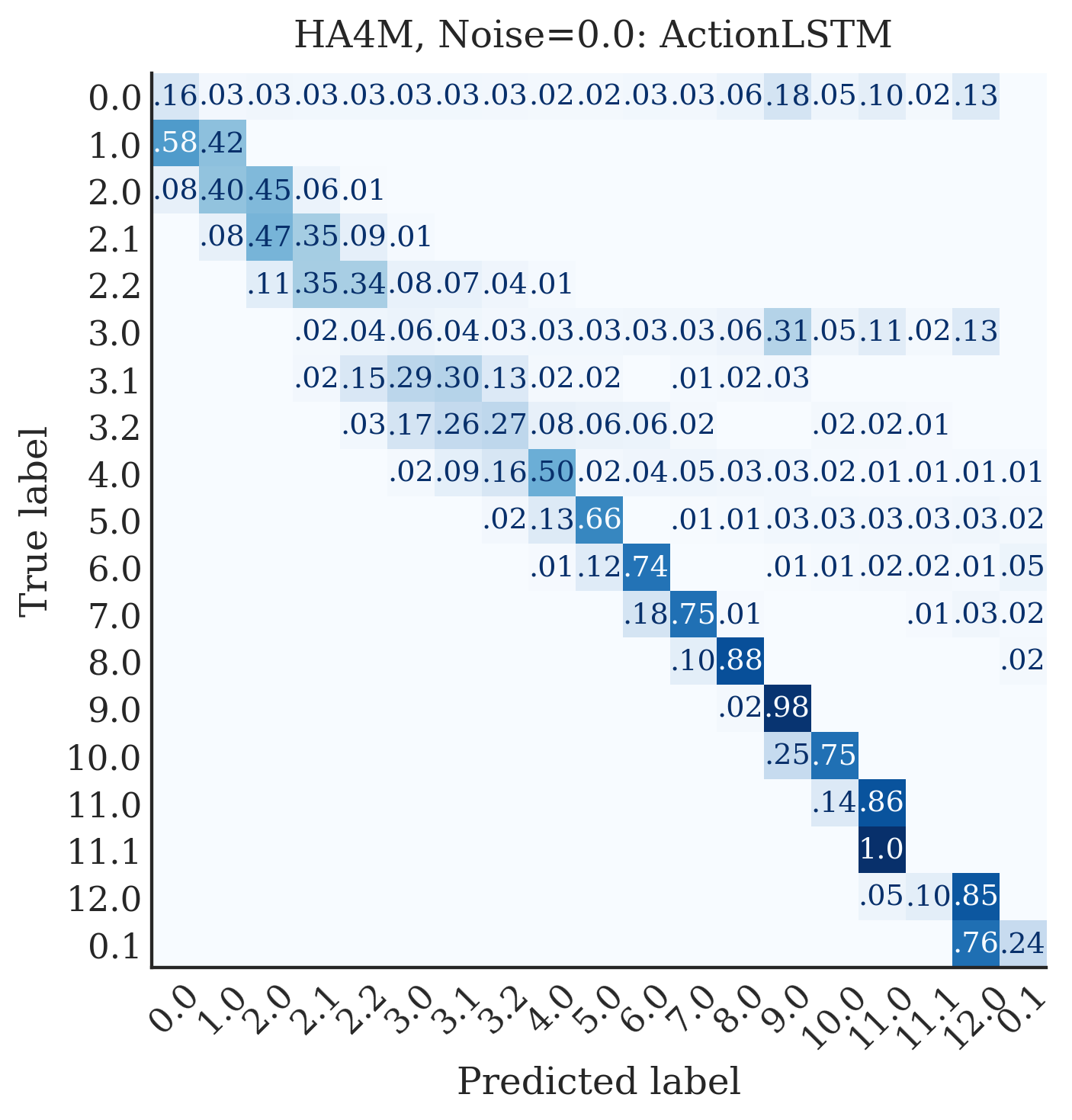}
    \end{subfigure}
    \hfill
    \vspace{-0.005em} 
    \caption{Confusion matrices for HA4M task state prediction with 0 noise added to input.}
    \label{fig:conf_mats_ha4m_nl0.0}
\end{figure*}

\begin{figure*}
    \centering
    \begin{subfigure}{\confmatsize\linewidth}
        \centering
        \includegraphics[width=\linewidth]{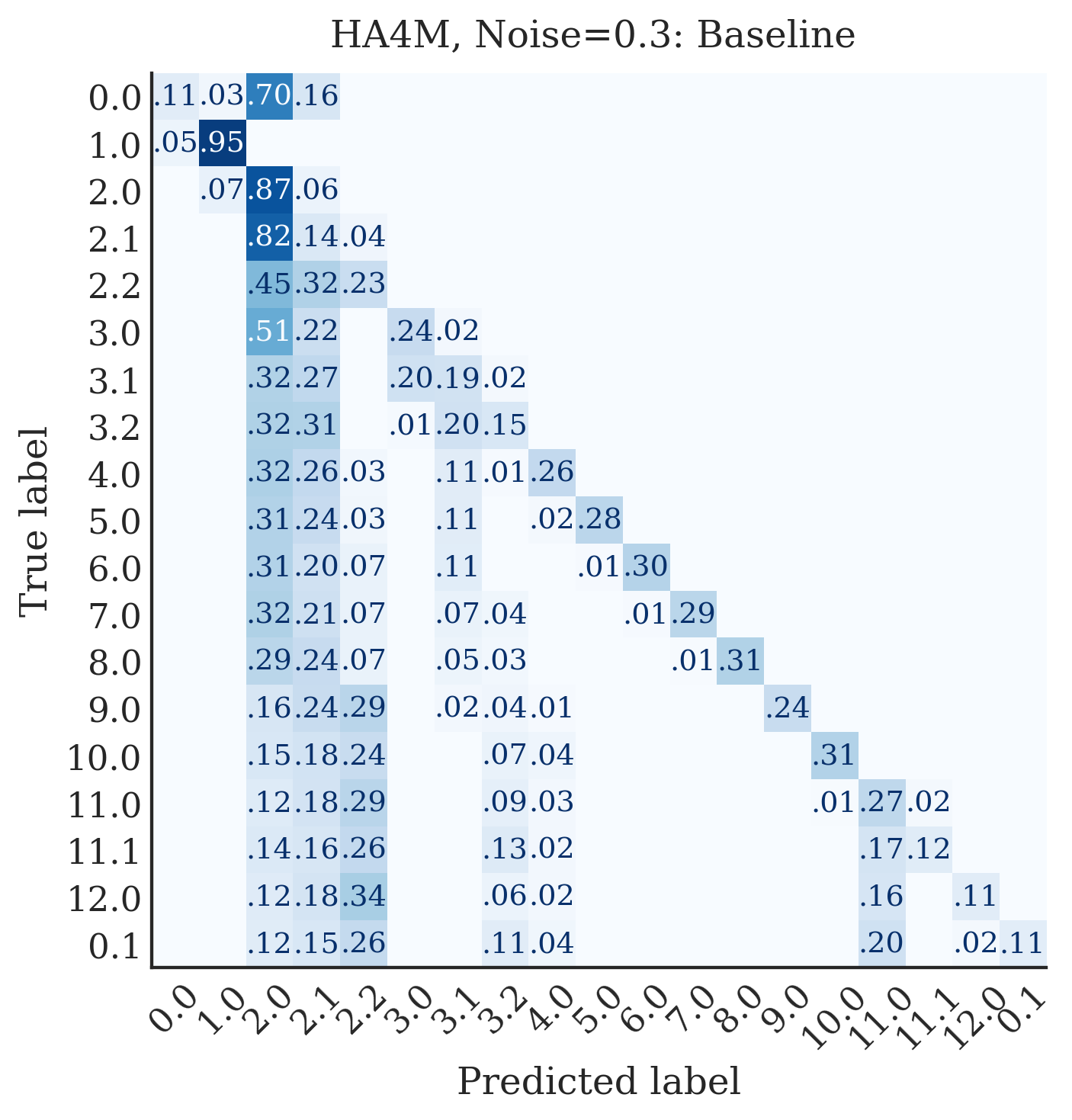}
    \end{subfigure}
    \hfill
    \begin{subfigure}{\confmatsize\linewidth}
        \centering
        \includegraphics[width=\linewidth]{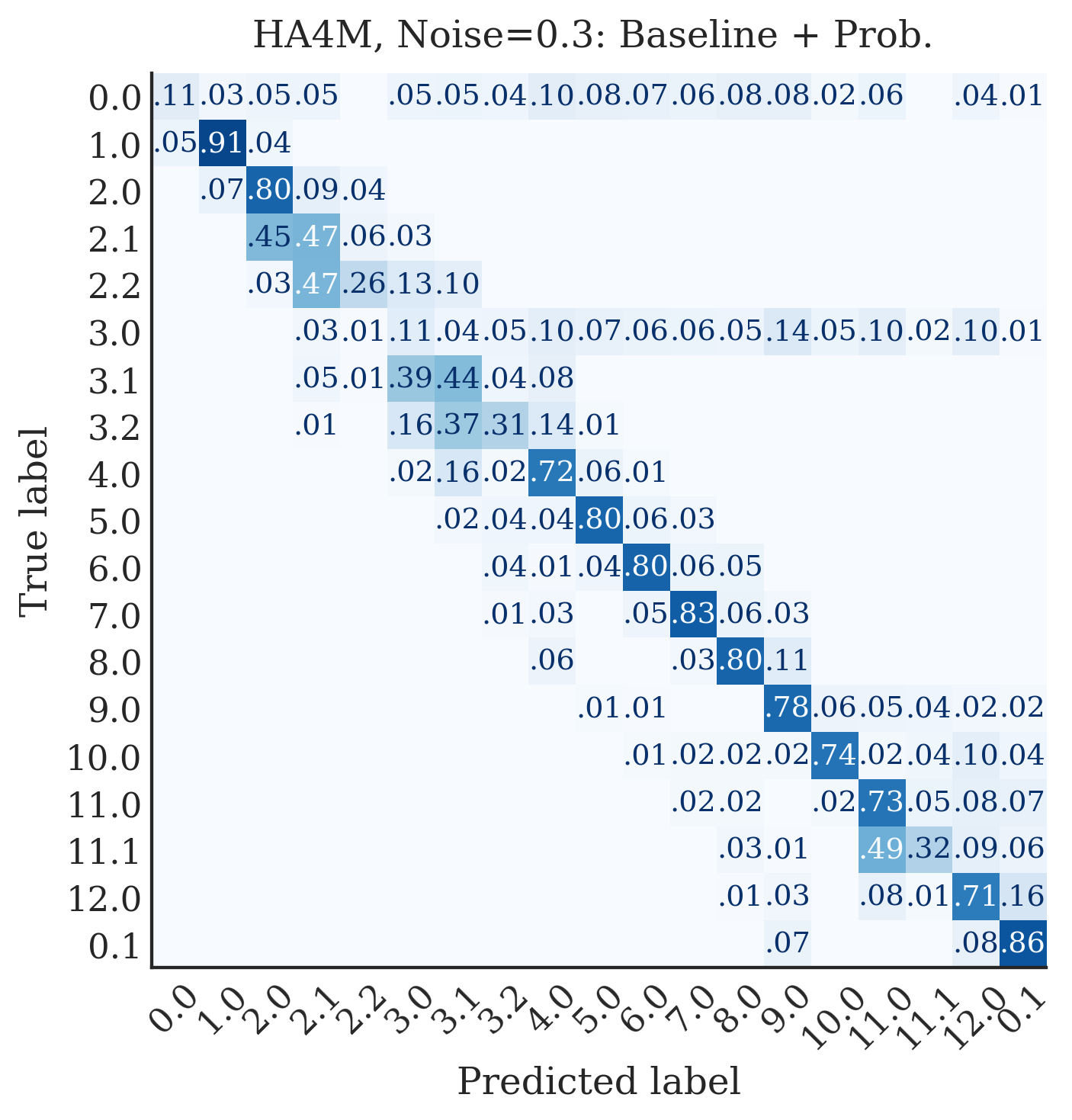}
    \end{subfigure}
    \hfill
    \begin{subfigure}{\confmatsize\linewidth}
        \centering
        \includegraphics[width=\linewidth]{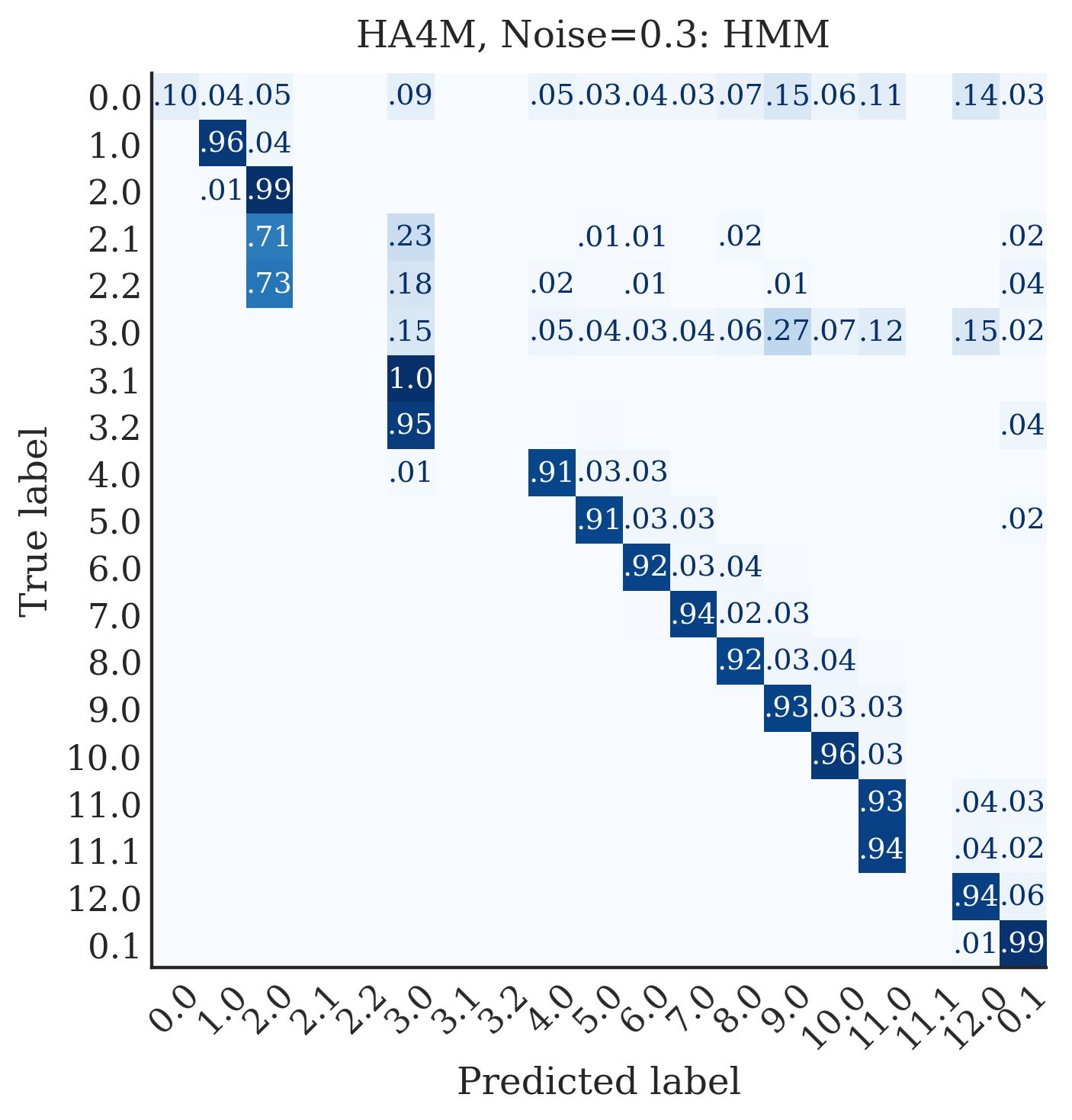}
    \end{subfigure}
    \hfill
    \begin{subfigure}{\confmatsize\linewidth}
        \centering
        \includegraphics[width=\linewidth]{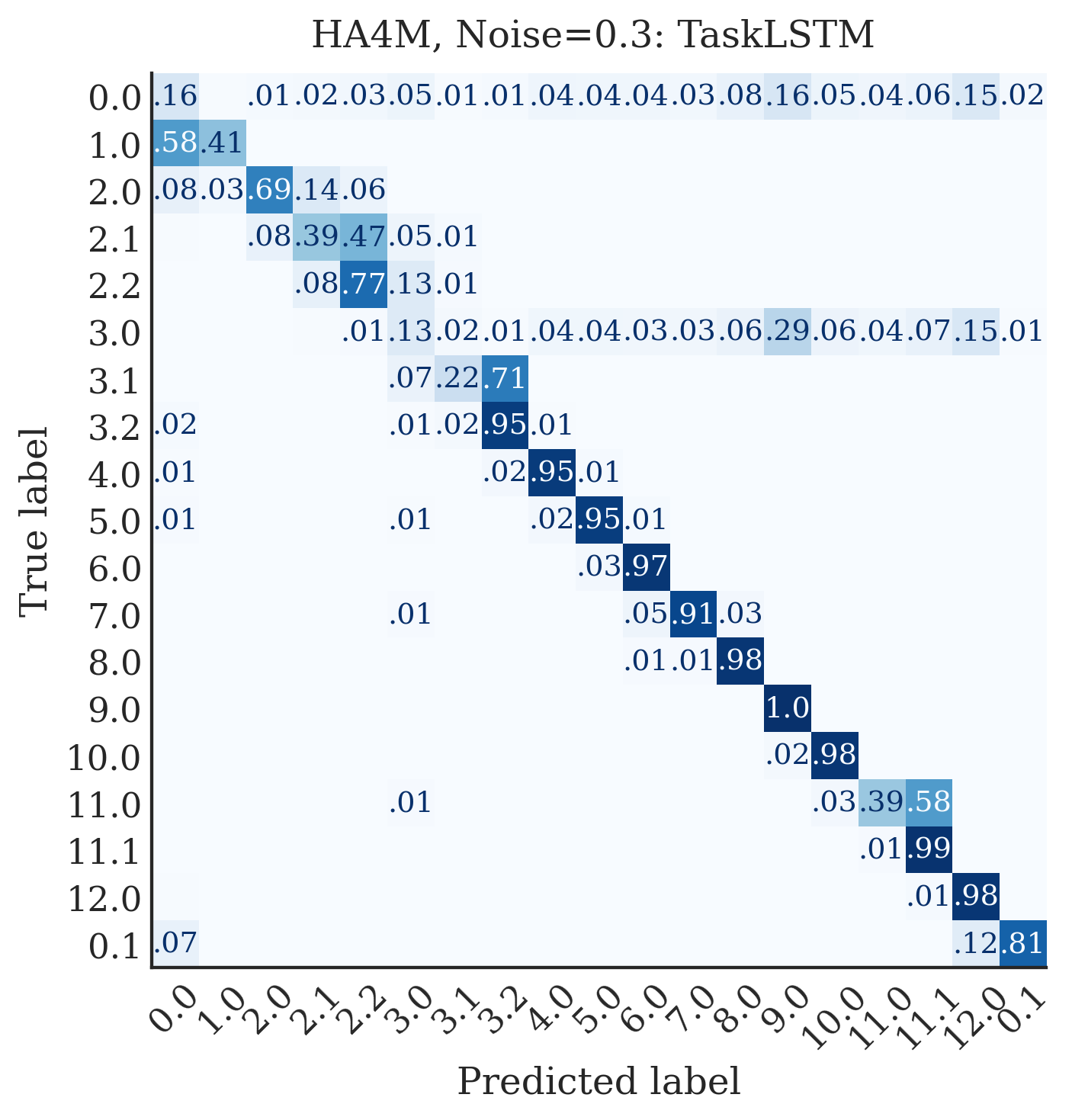}
    \end{subfigure}
    \hfill
    \begin{subfigure}{\confmatsize\linewidth}
        \centering
        \includegraphics[width=\linewidth]{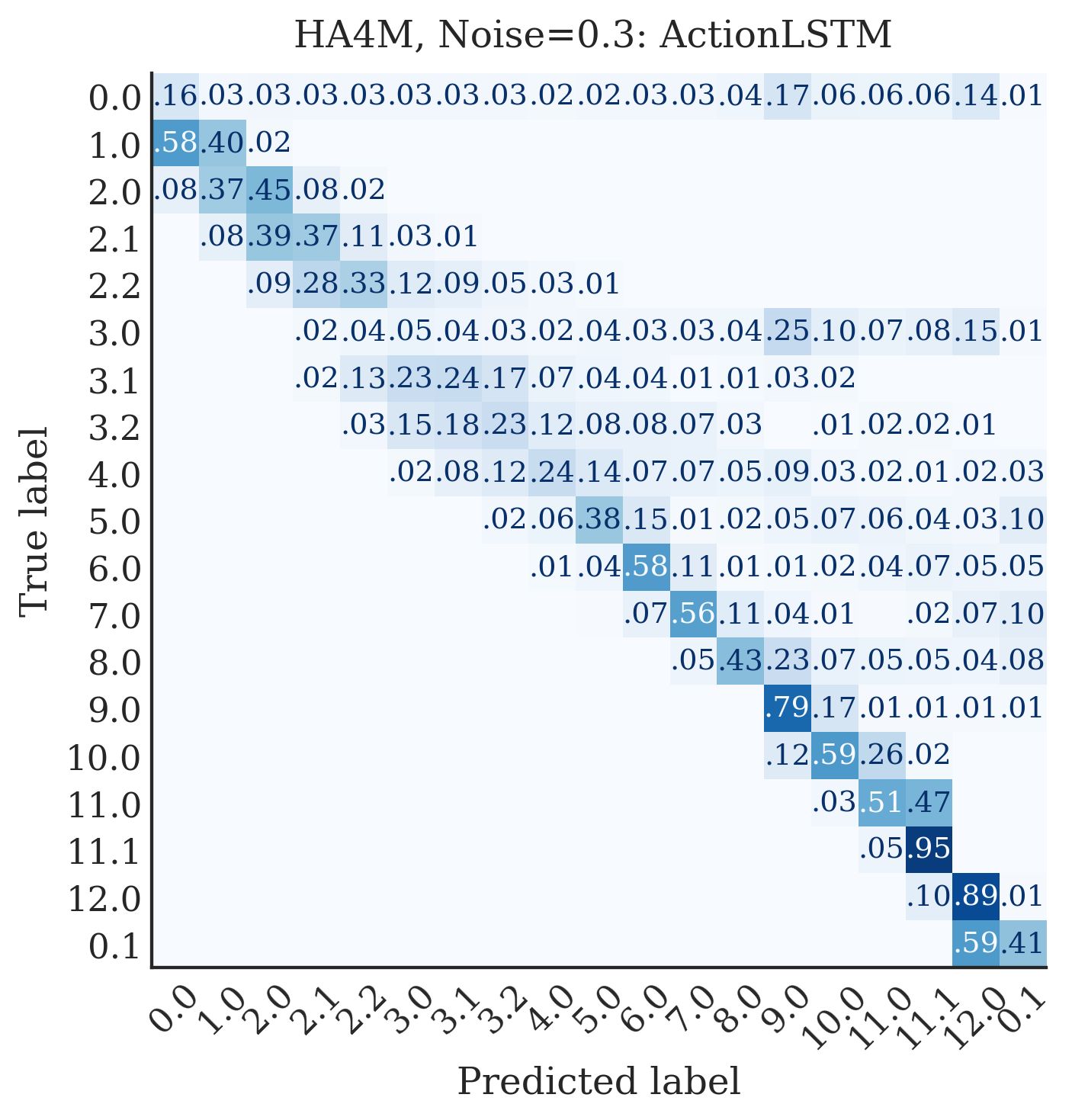}
    \end{subfigure}
    \hfill
    \vspace{-0.005em} 
    \caption{Confusion matrices for HA4M task state prediction with 0.3 noise added to input.}
    \label{fig:conf_mats_ha4m_nl0.3}
\end{figure*}
\begin{figure*}
    \centering
    \begin{subfigure}{\confmatsize\linewidth}
        \centering
        \includegraphics[width=\linewidth]{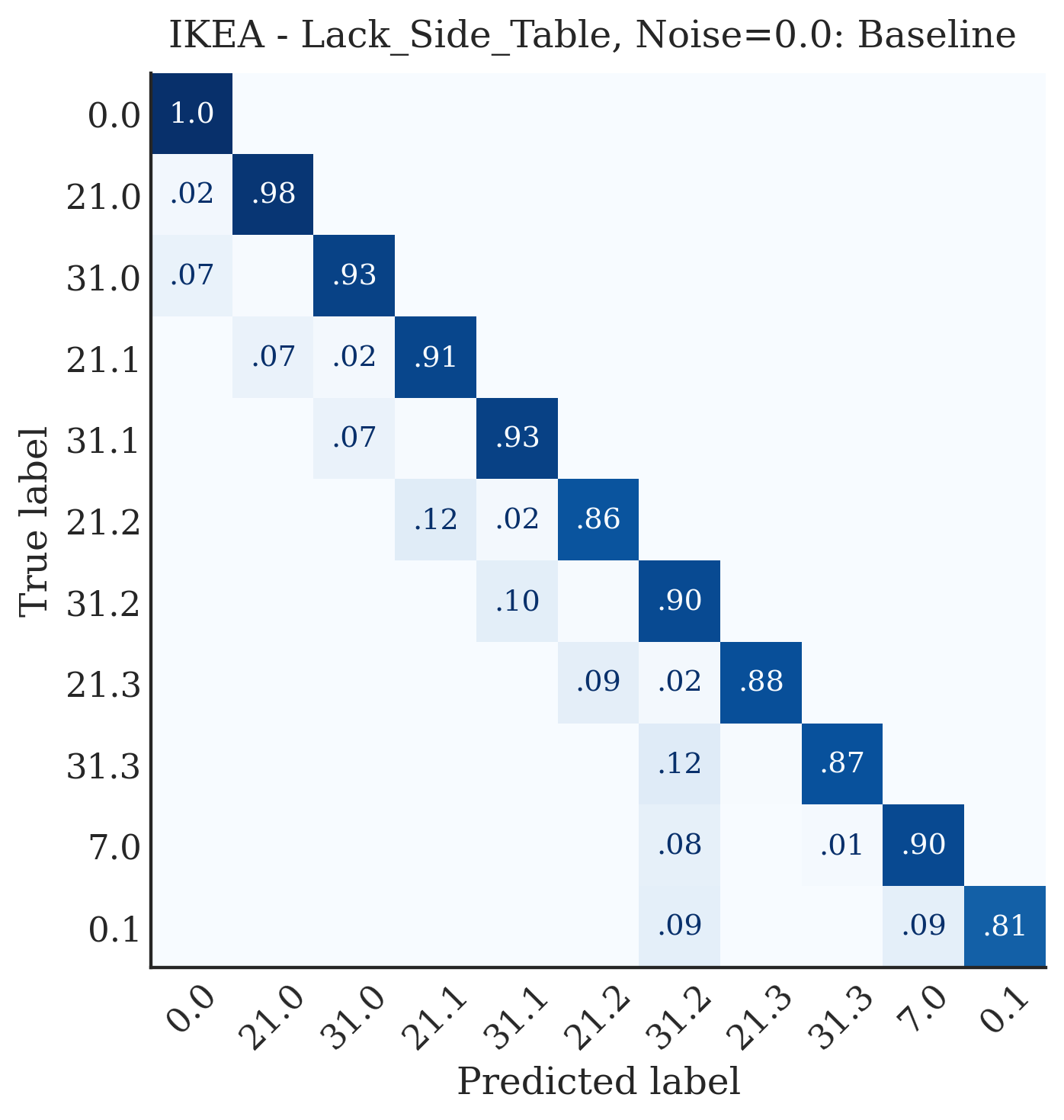}
    \end{subfigure}
    \hfill
    \begin{subfigure}{\confmatsize\linewidth}
        \centering
        \includegraphics[trim={2mm 0mm 4mm 0mm},clip,width=\linewidth]{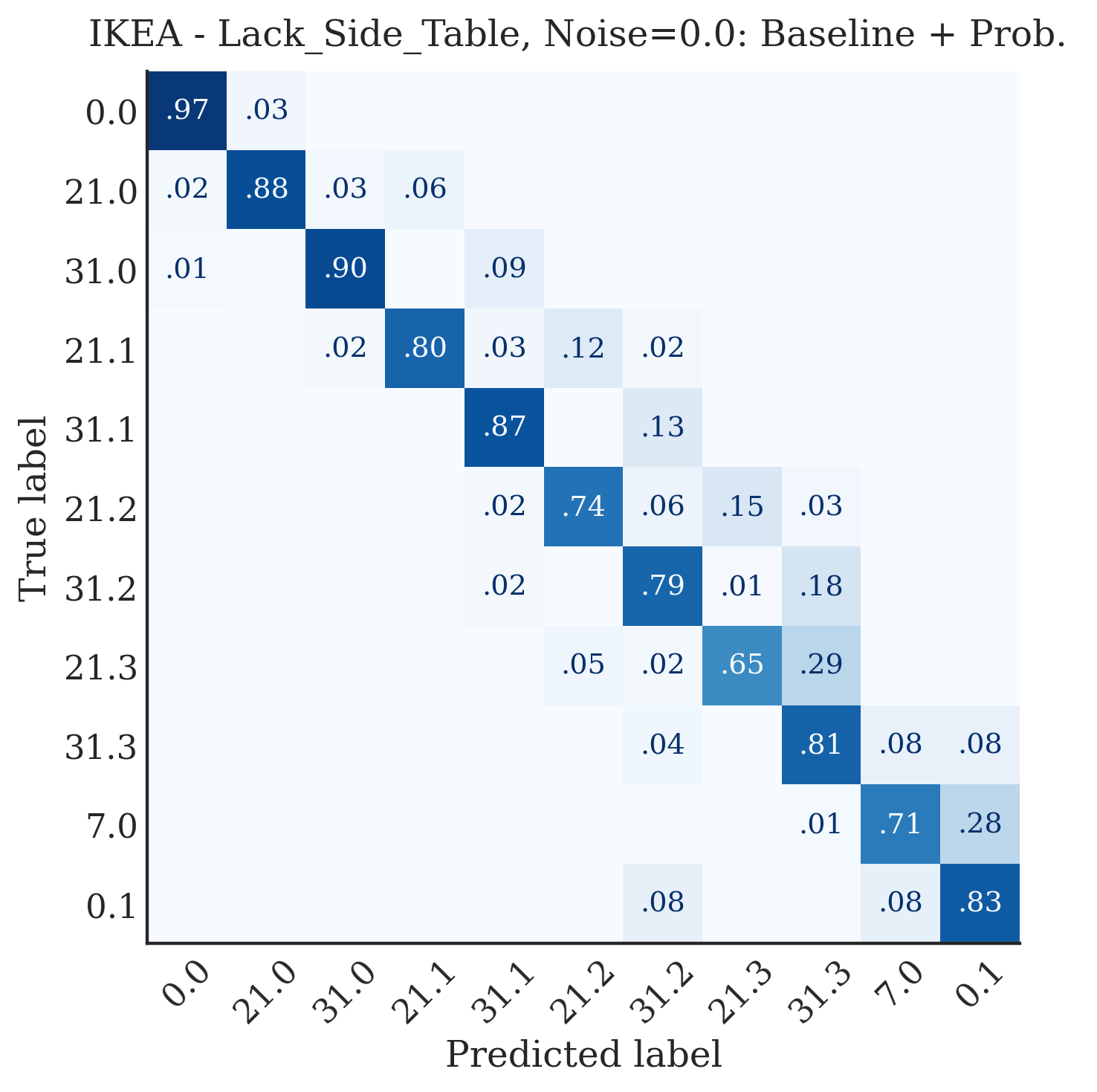}
    \end{subfigure}
    \hfill
    \begin{subfigure}{\confmatsize\linewidth}
        \centering
        \includegraphics[width=\linewidth]{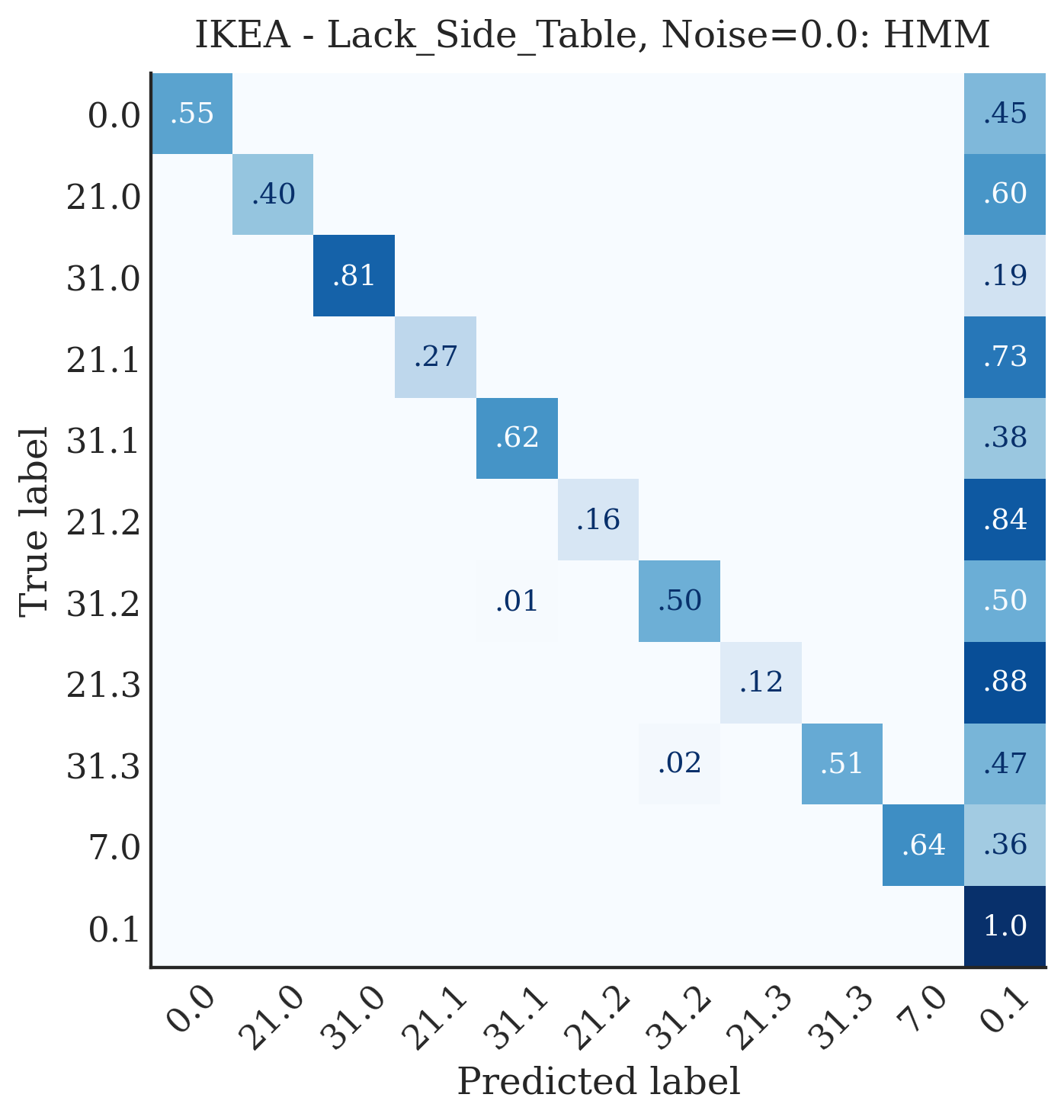}
    \end{subfigure}
    \vspace{0.005em} 
    \hfill
    \begin{subfigure}{\confmatsize\linewidth}
        \centering
        \includegraphics[width=\linewidth]{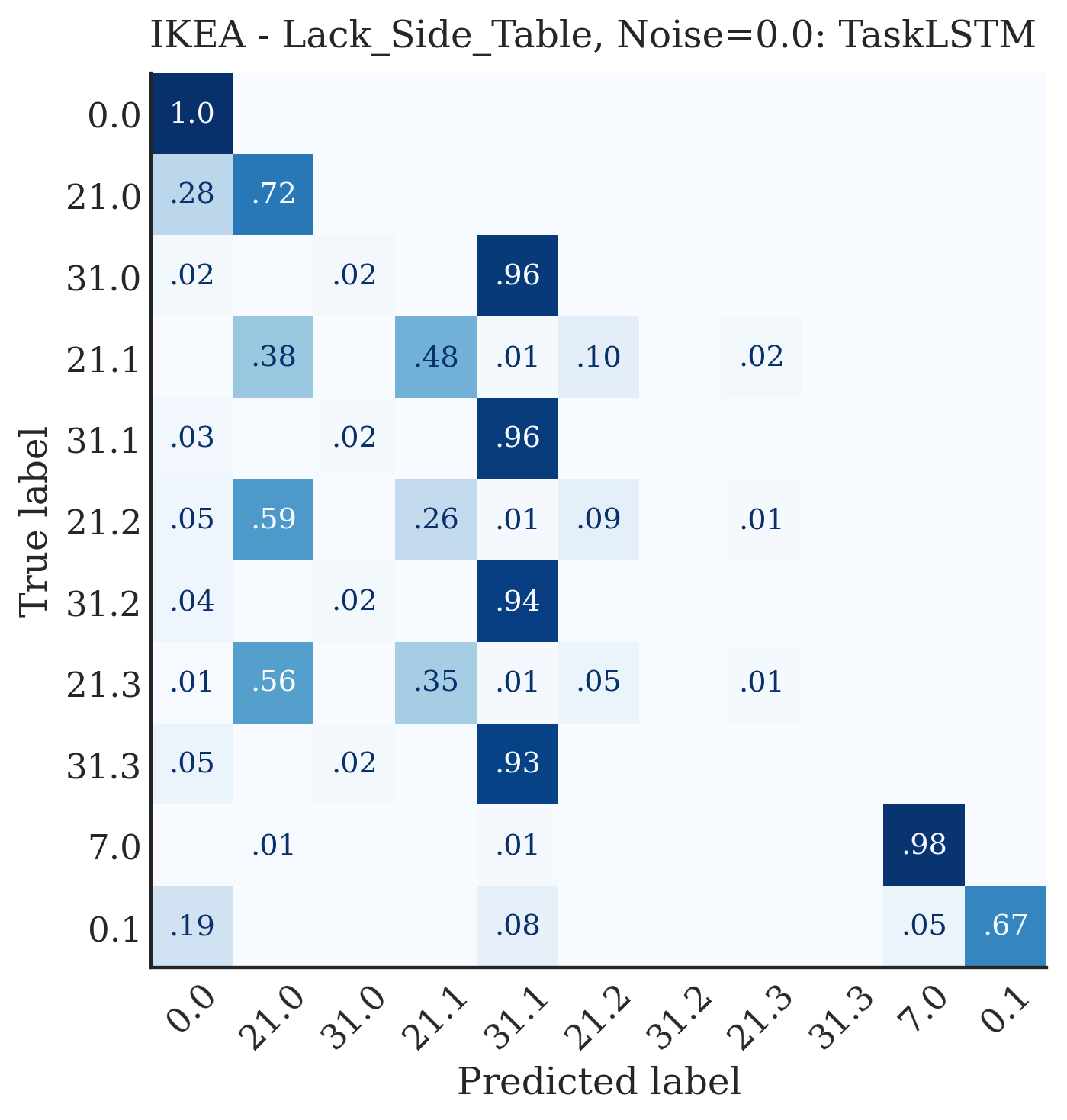}
    \end{subfigure}
    \hfill
    \begin{subfigure}{\confmatsize\linewidth}
        \centering
        \includegraphics[trim={2mm 0mm 2mm 0mm},clip,width=\linewidth]{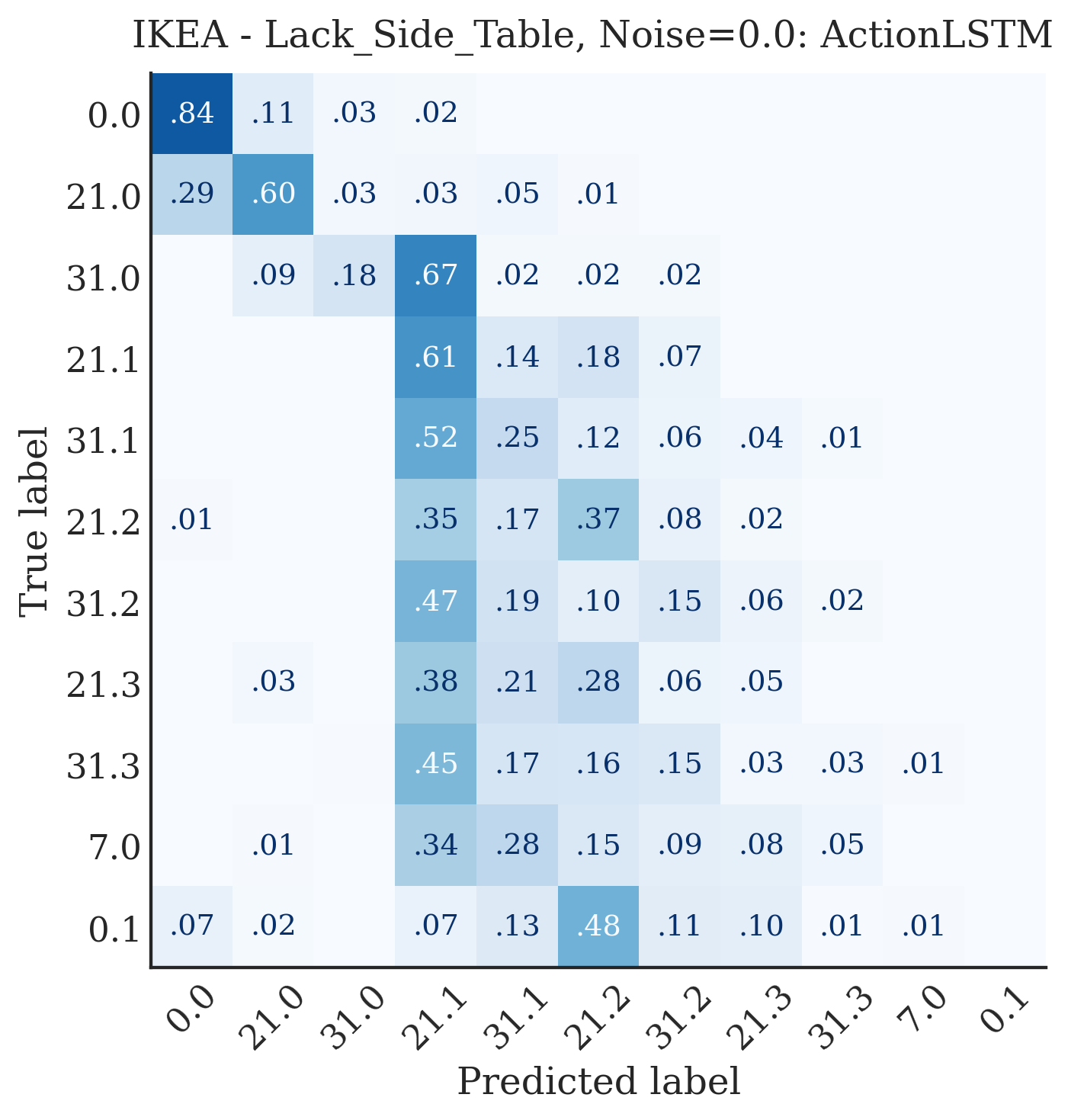}
    \end{subfigure}
    \hfill
    \vspace{-0.005em} 
    \caption{Confusion matrices for IKEA Lack Side Table task state prediction with 0 noise added to input.}
    \label{fig:conf_mats_Lack_Side_Table_nl0.0}
\end{figure*}

\begin{figure*}
    \centering
    \begin{subfigure}{\confmatsize\linewidth}
        \centering
        \includegraphics[width=\linewidth]{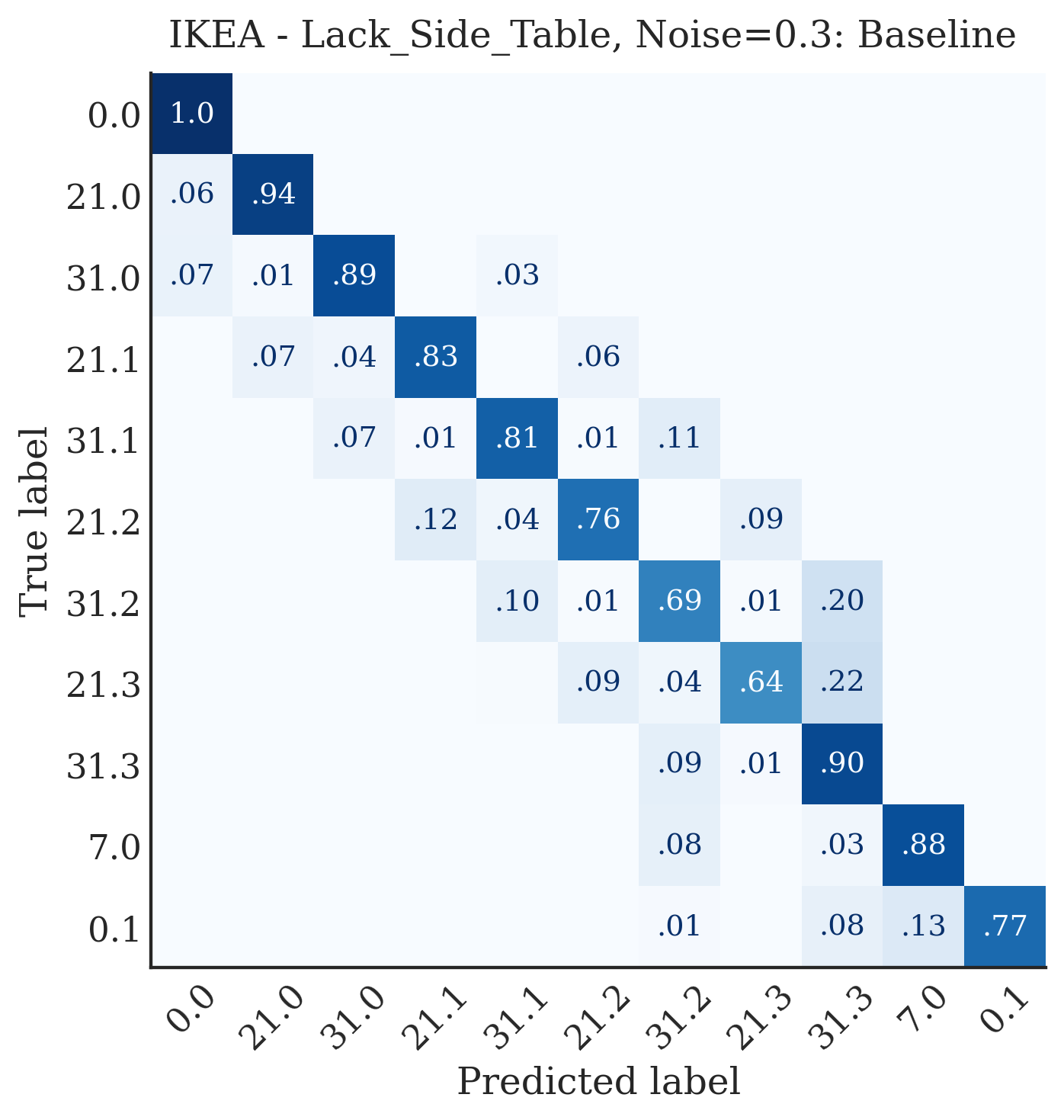}
    \end{subfigure}
    \hfill
    \begin{subfigure}{\confmatsize\linewidth}
        \centering
        \includegraphics[trim={2mm 0mm 4mm 0mm},clip,width=\linewidth]{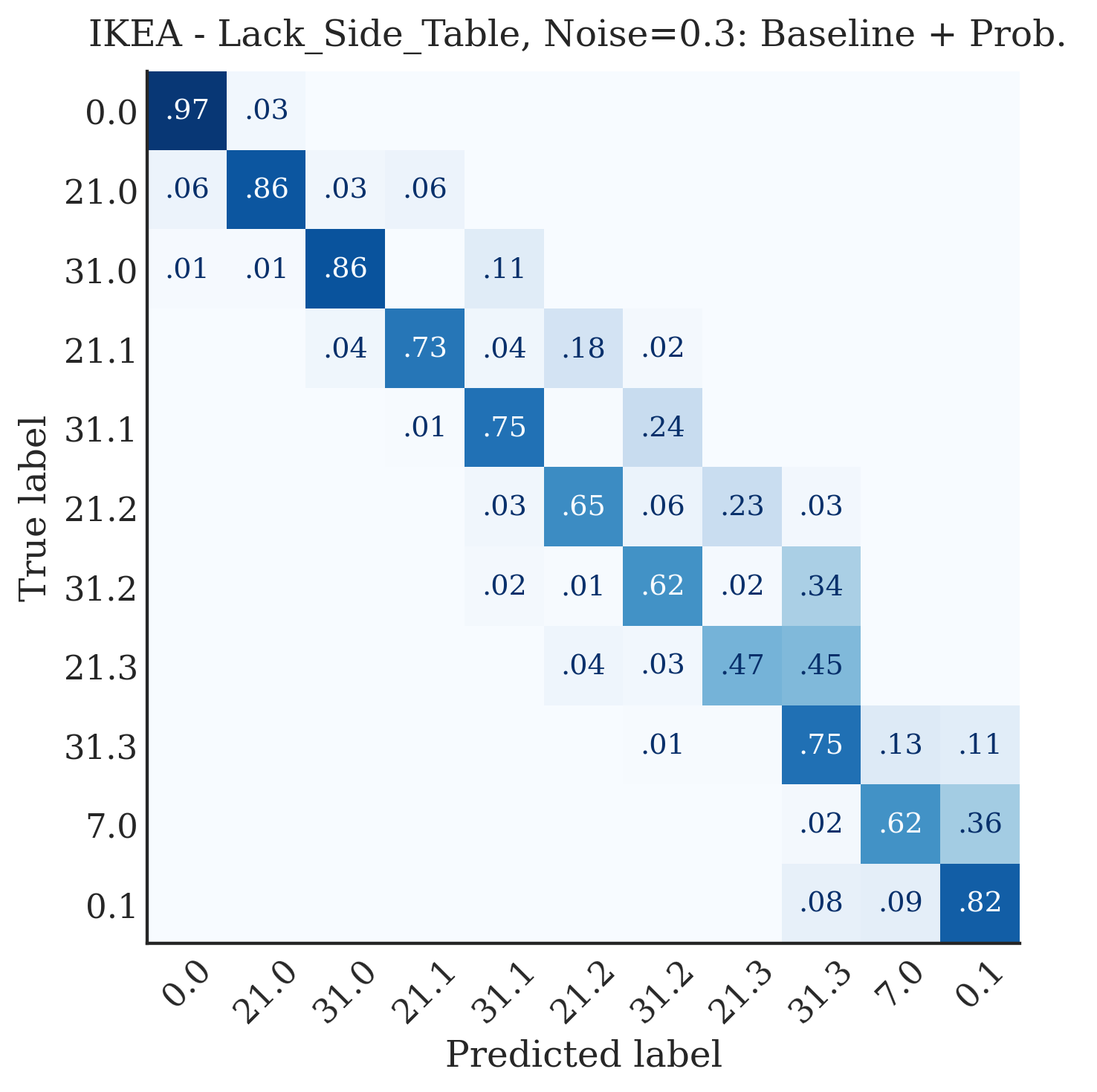}
    \end{subfigure}
    \hfill
    \begin{subfigure}{\confmatsize\linewidth}
        \centering
        \includegraphics[width=\linewidth]{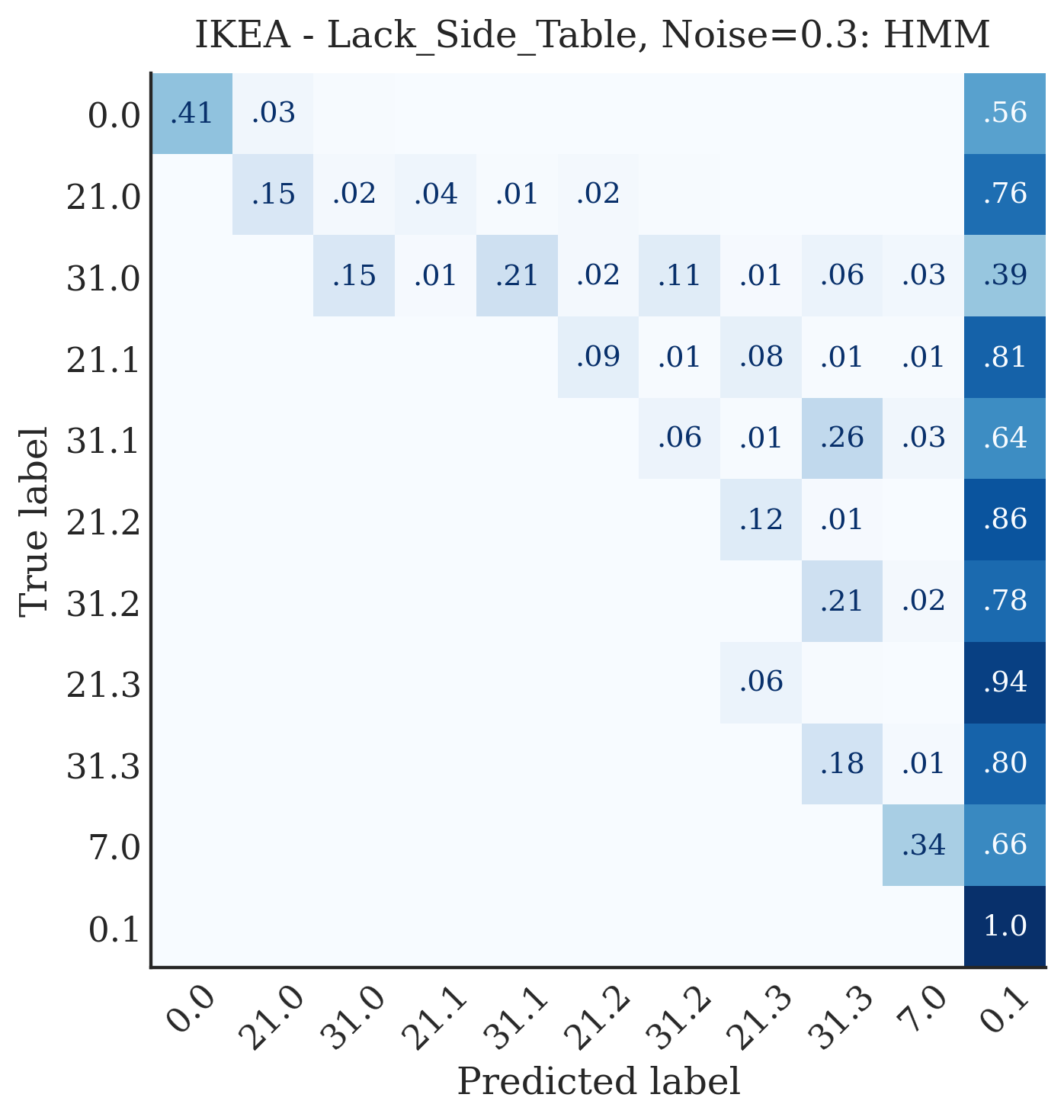}
    \end{subfigure}
    \hfill
    \begin{subfigure}{\confmatsize\linewidth}
        \centering
        \includegraphics[width=\linewidth]{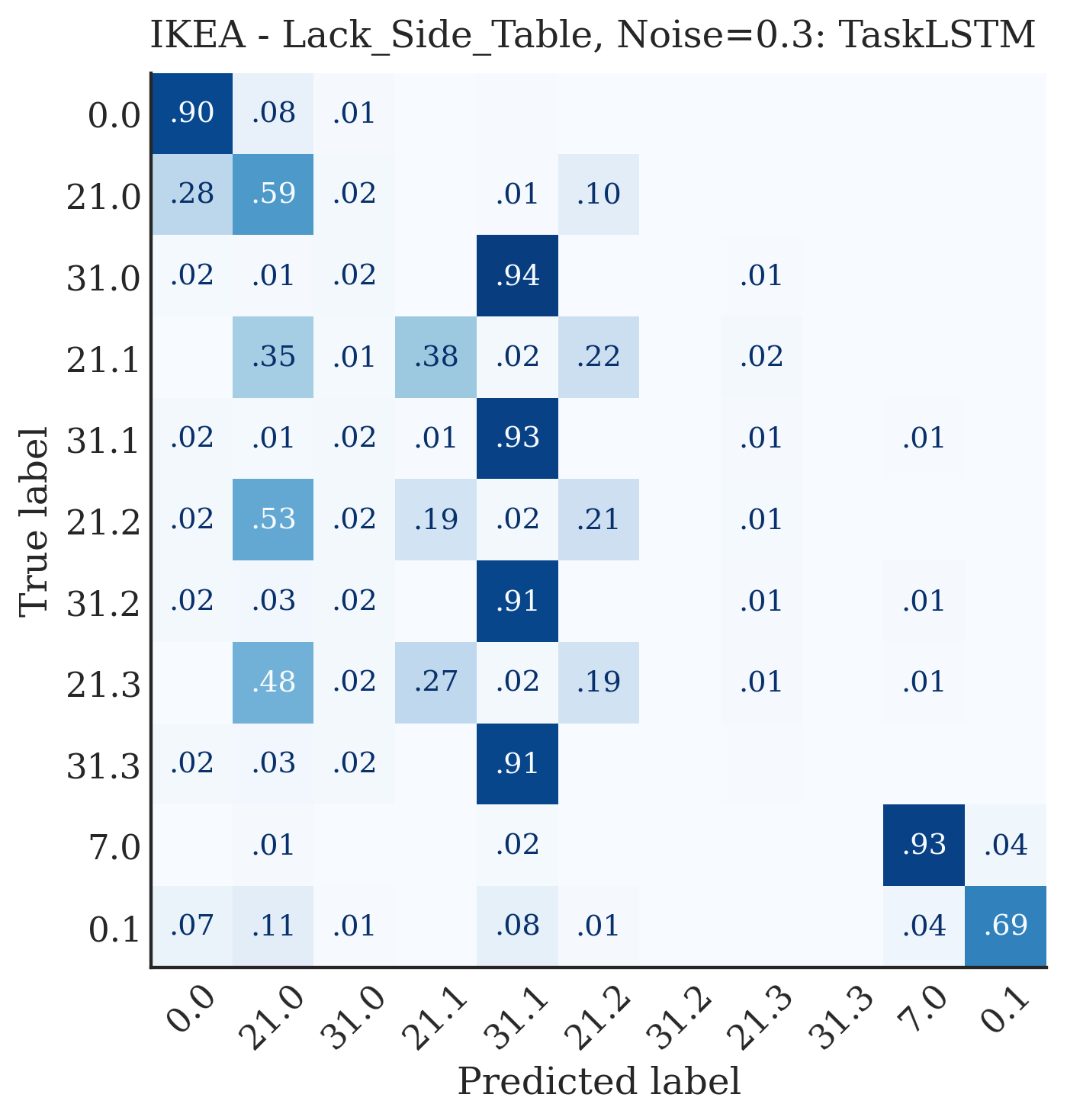}
    \end{subfigure}
    \hfill
    \begin{subfigure}{\confmatsize\linewidth}
        \centering
        \includegraphics[trim={2mm 0mm 2mm 0mm},clip,width=\linewidth]{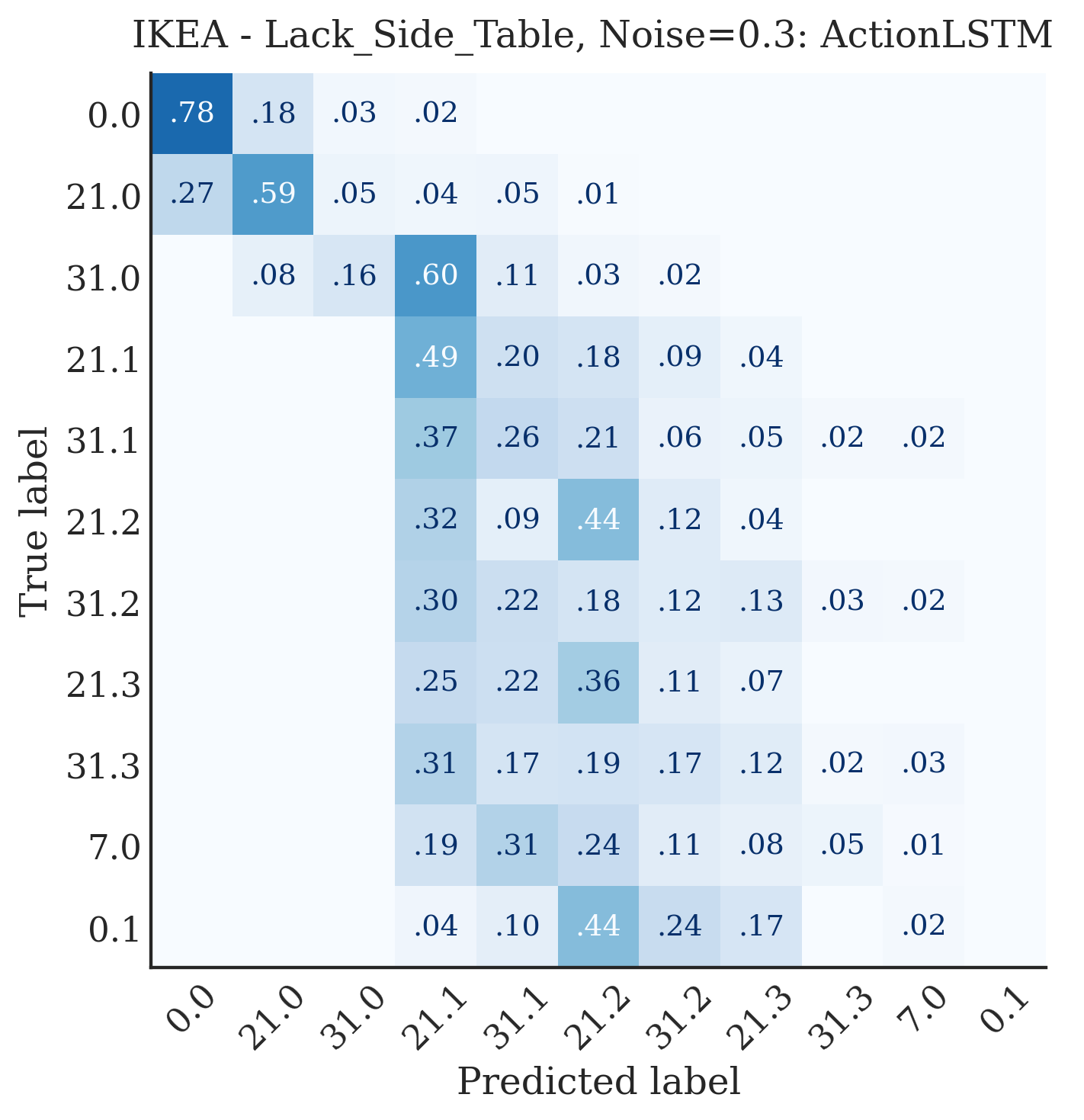}
    \end{subfigure}
    \hfill
    \vspace{-0.005em} 
    \caption{Confusion matrices for IKEA Lack Side Table task state prediction with 0.3 noise added to input.}
    \label{fig:conf_mats_Lack_Side_Table_nl0.3}
    \vspace{-3mm}
\end{figure*}

\subsection{Action Recognition Experiments}\label{sec:har_data}
The second stage of experiments uses a HAR input to provide a realistic input to the models. An ST-GCN model is trained on the skeleton data associated with each of the datasets to output the action prediction at each timestep. The ST-GCN architecture is common for skeleton-based action recognition and, while more advanced methods are available, has shown good results in previous research and is easily deployable~\cite{xiao_intelligent_2025, soleymani2025skeleton}. The model implementation is that made available by Yan et al.~\cite{yan2018spatial}. Two models are trained, one for each dataset, with the model trained on the training data for each dataset. Both are trained for 50 epochs with a 3\,s input sequence, batch size of 64 and 0.1 base learning rate. The HA4M dataset uses the Azure Kinect skeleton configuration and outputs the 13 task actions, while the IKEA classifier uses the OpenPose skeleton configuration and outputs the 33 possible actions. The HA4M model achieves an accuracy of 0.89 and F1 score of 0.88, while the IKEA model has an accuracy of 0.67 and F1 score of 0.27. The low scores for the IKEA dataset are in line with previous skeleton-based action recognition on the dataset~\cite{ben2021ikea}, and the imperfect results on both datasets help demonstrate the state prediction with suboptimal inputs.

\begin{figure*}
    \centering
    \begin{subfigure}{\confmatsize\linewidth}
        \centering
        \includegraphics[width=\linewidth]{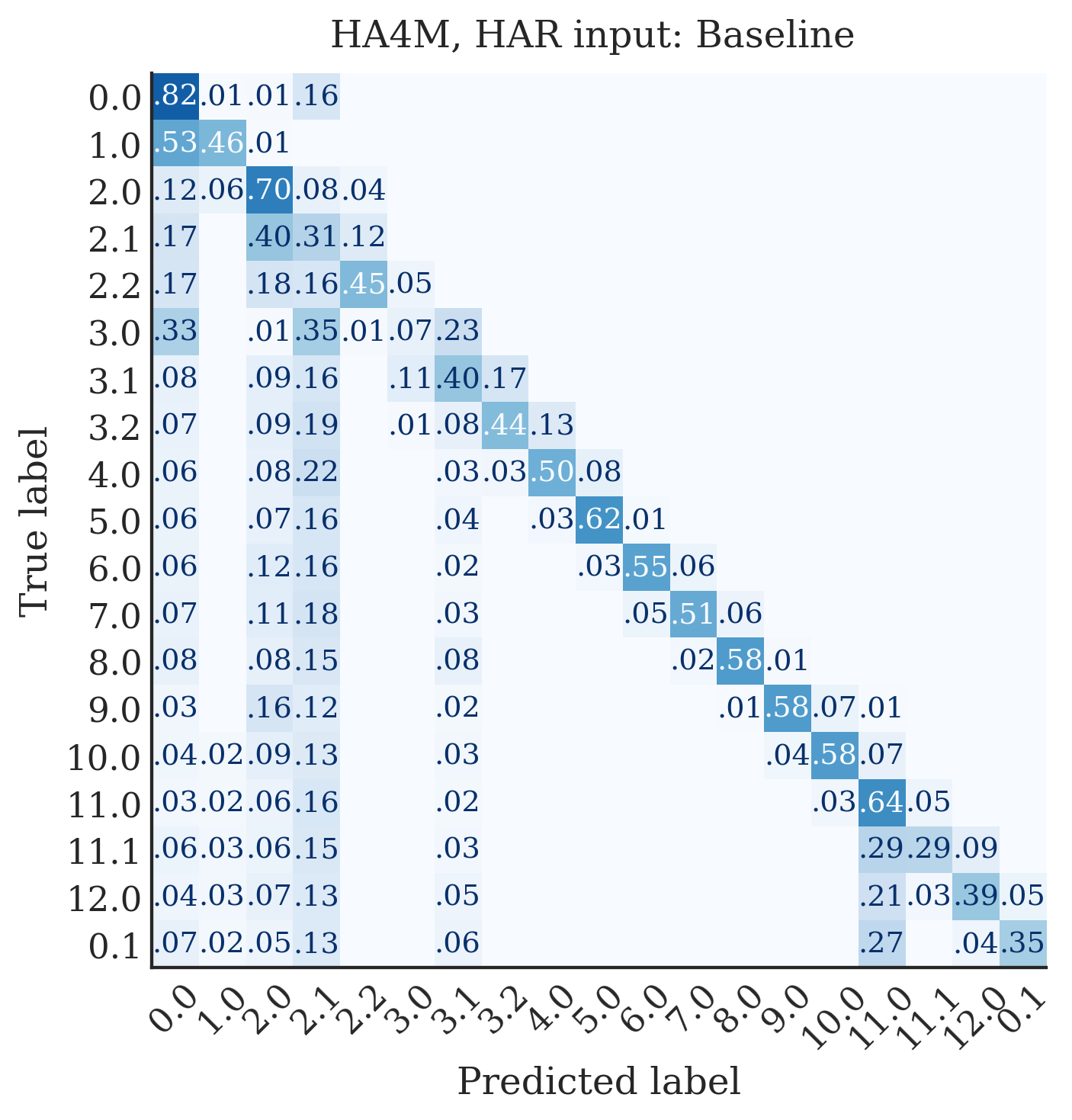}
    \end{subfigure}
    \hfill
    \begin{subfigure}{\confmatsize\linewidth}
        \centering
        \includegraphics[width=\linewidth]{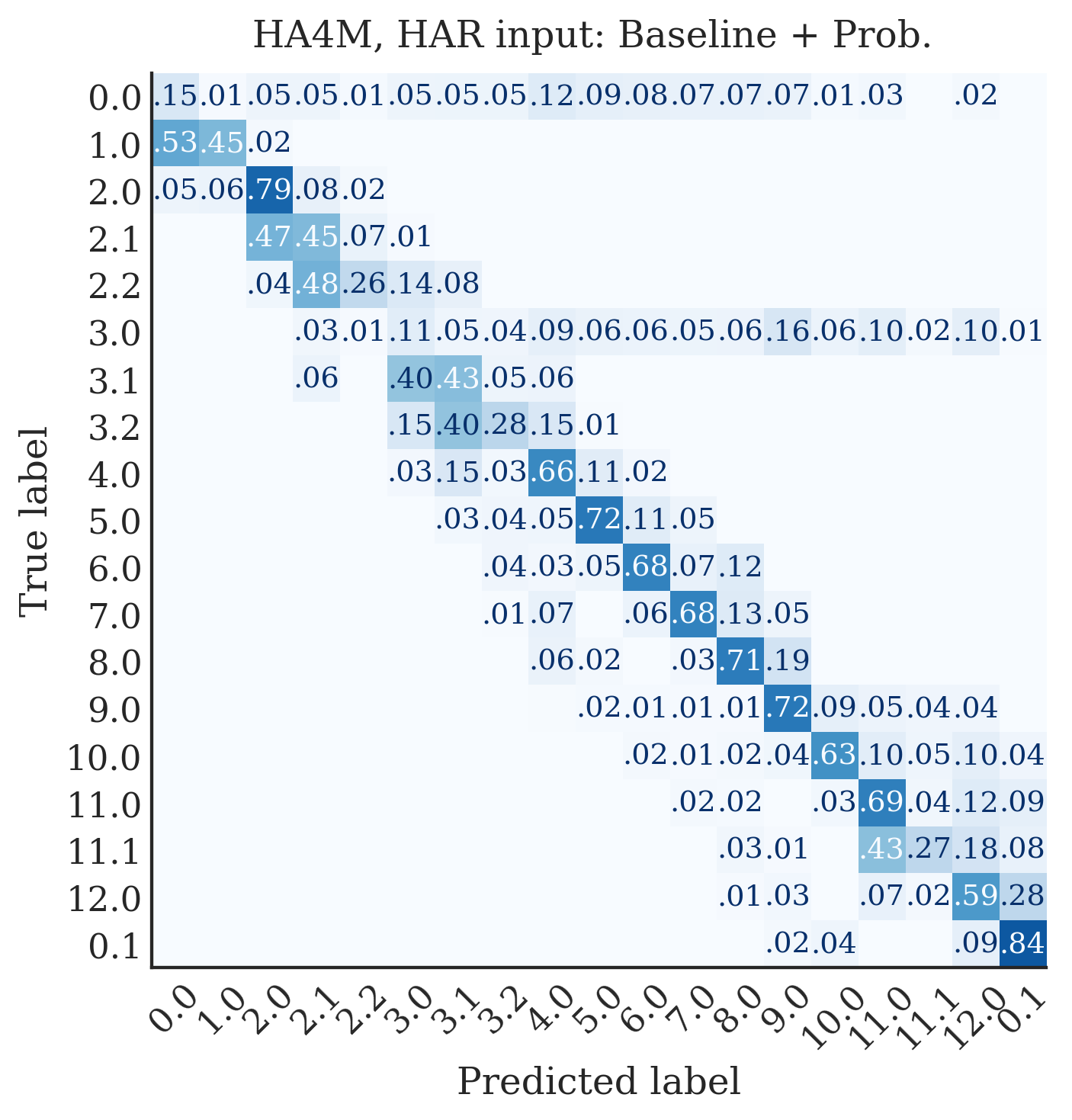}
    \end{subfigure}
    \hfill
    \begin{subfigure}{\confmatsize\linewidth}
        \centering
        \includegraphics[width=\linewidth]{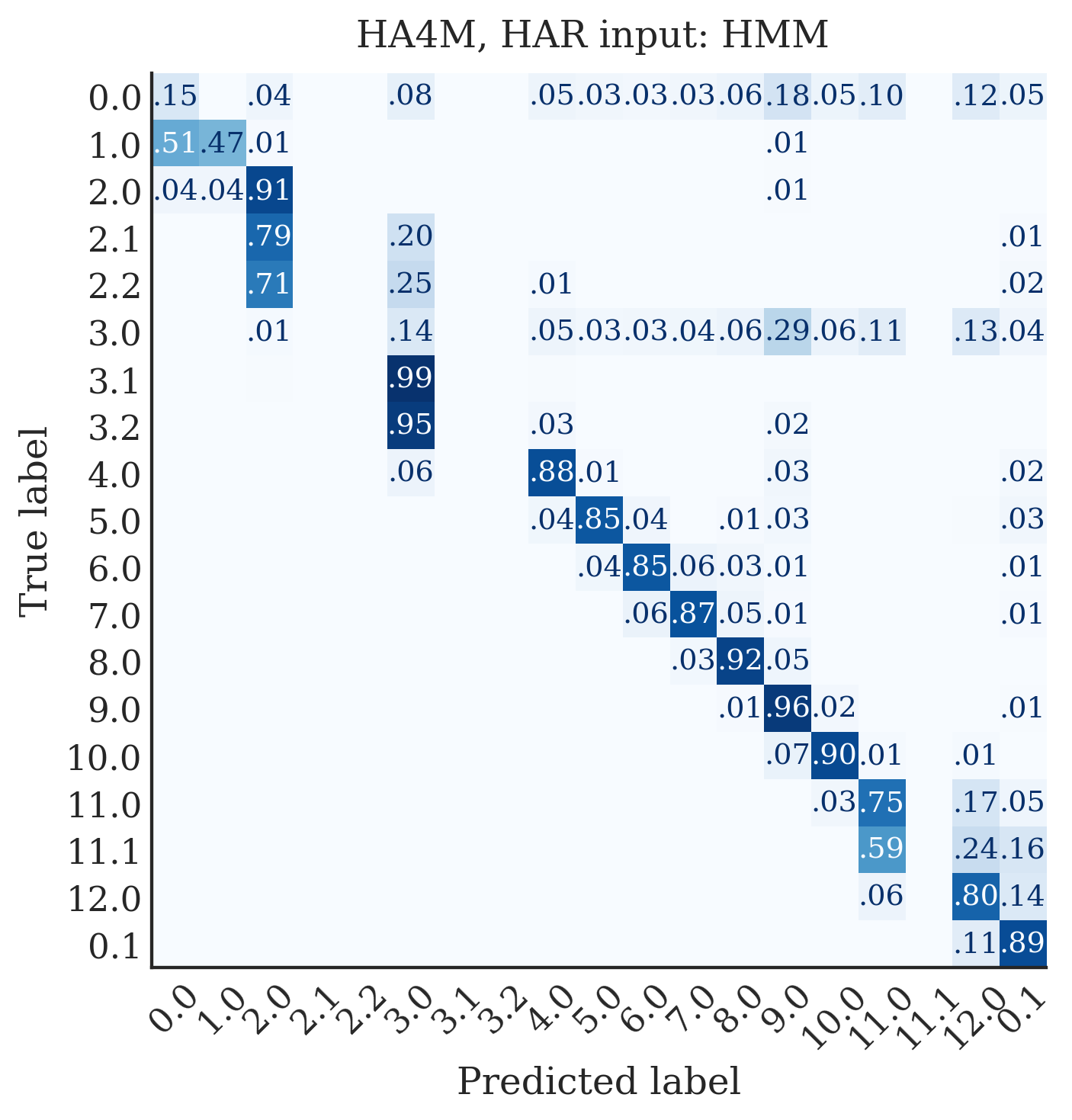}
    \end{subfigure}
    \hfill
    \begin{subfigure}{\confmatsize\linewidth}
        \centering
        \includegraphics[width=\linewidth]{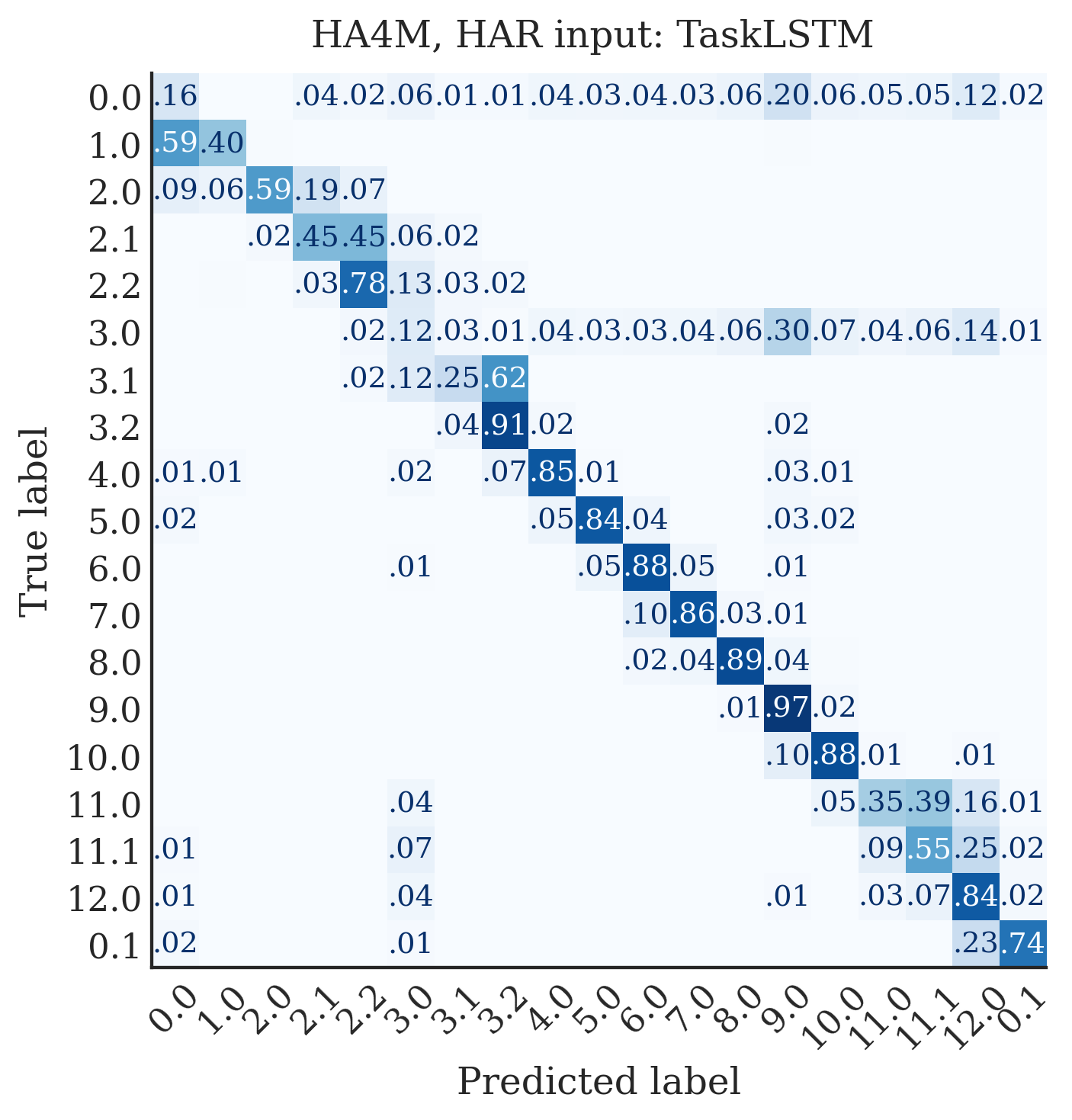}
    \end{subfigure}
    \hfill
    \begin{subfigure}{\confmatsize\linewidth}
        \centering
        \includegraphics[width=\linewidth]{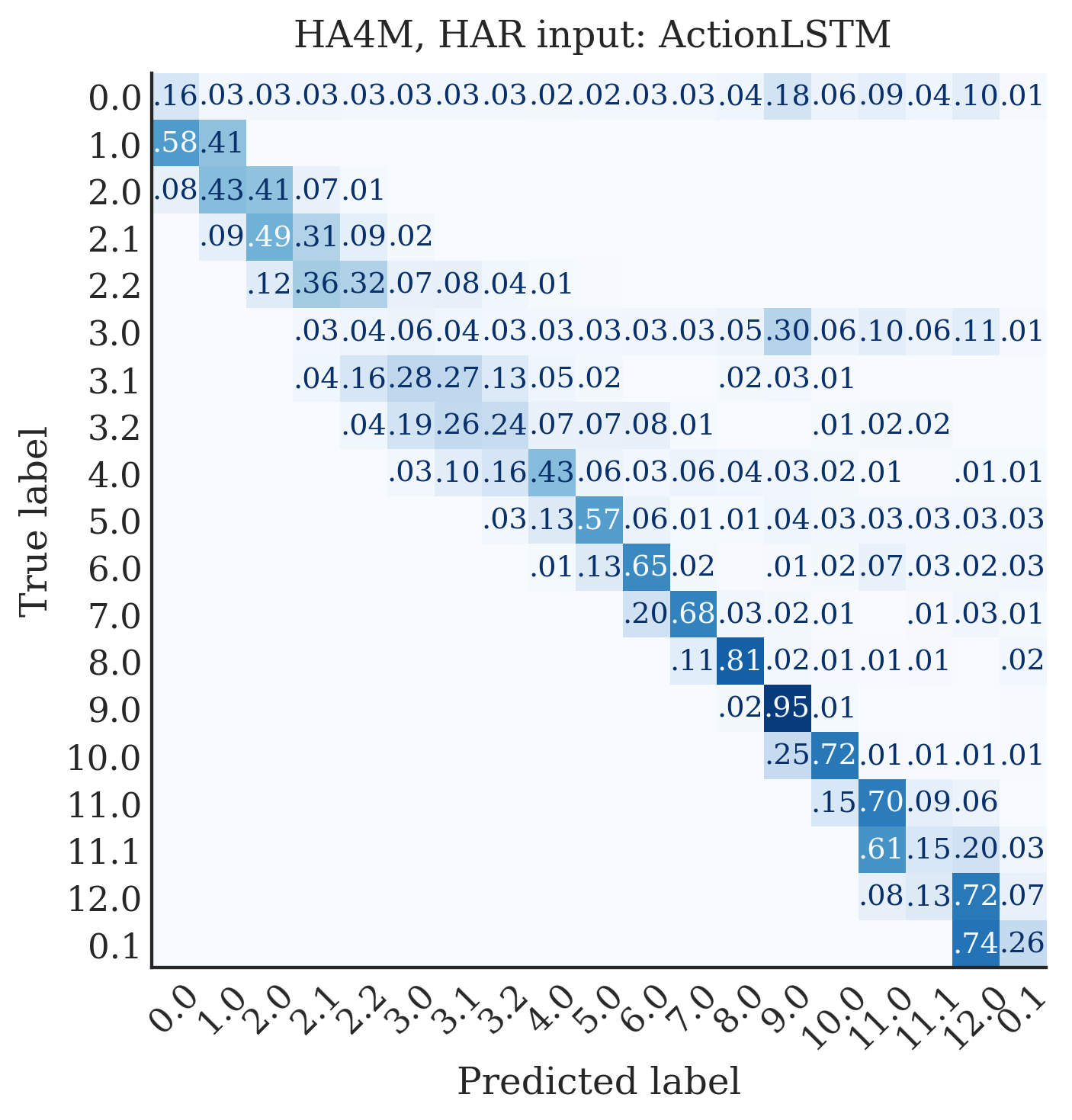}
    \end{subfigure}
    \hfill
    \vspace{-0.005em} 
    \caption{Confusion matrices for HA4M task state prediction with HAR input.}
    \label{fig:conf_mats_ha4m_har}
\end{figure*}
\begin{figure*}
    \centering
    \begin{subfigure}{\confmatsize\linewidth}
        \centering
        \includegraphics[width=\linewidth]{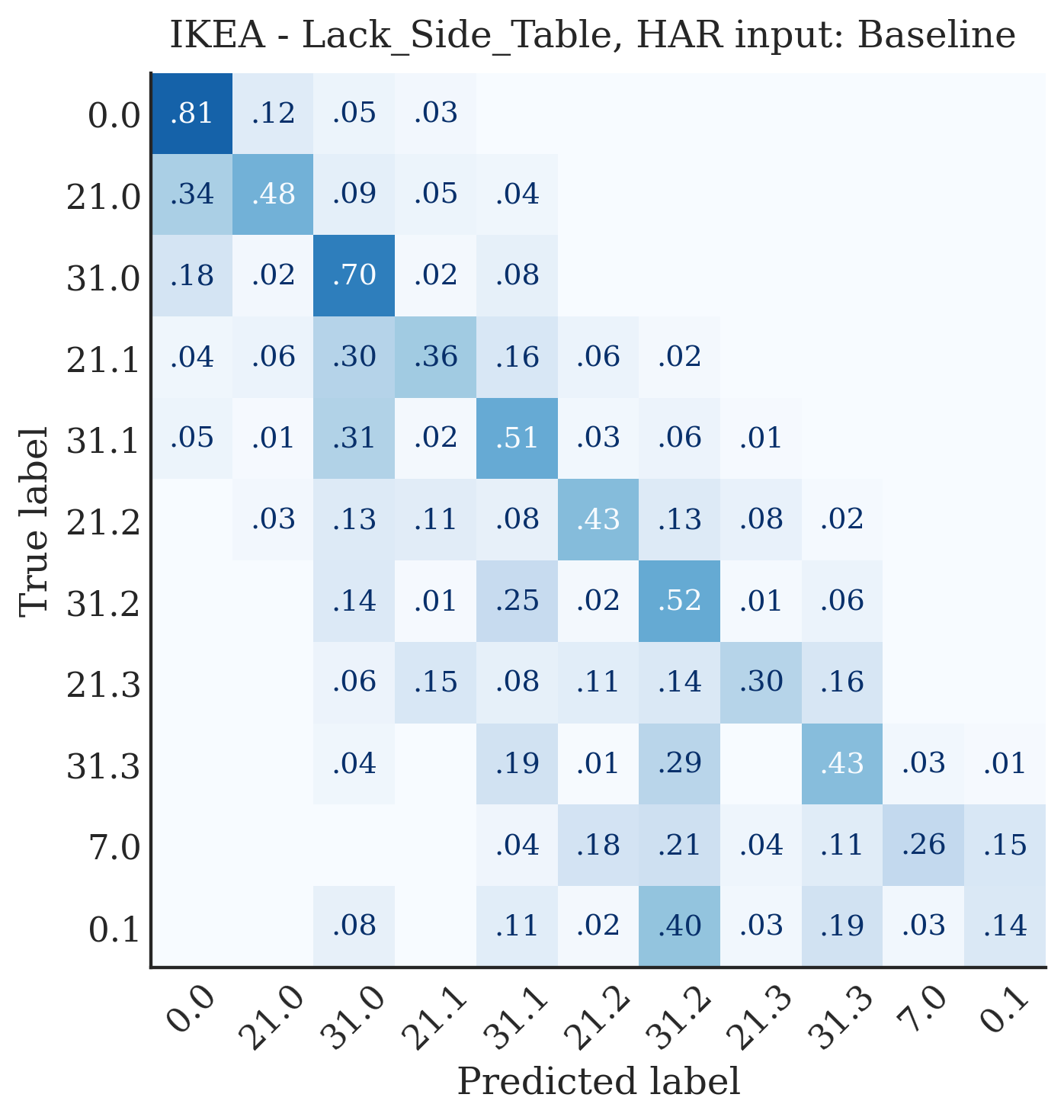}
    \end{subfigure}
    \hfill
    \begin{subfigure}{\confmatsize\linewidth}
        \centering
        \includegraphics[trim={2mm 0mm 4mm 0mm},clip,width=\linewidth]{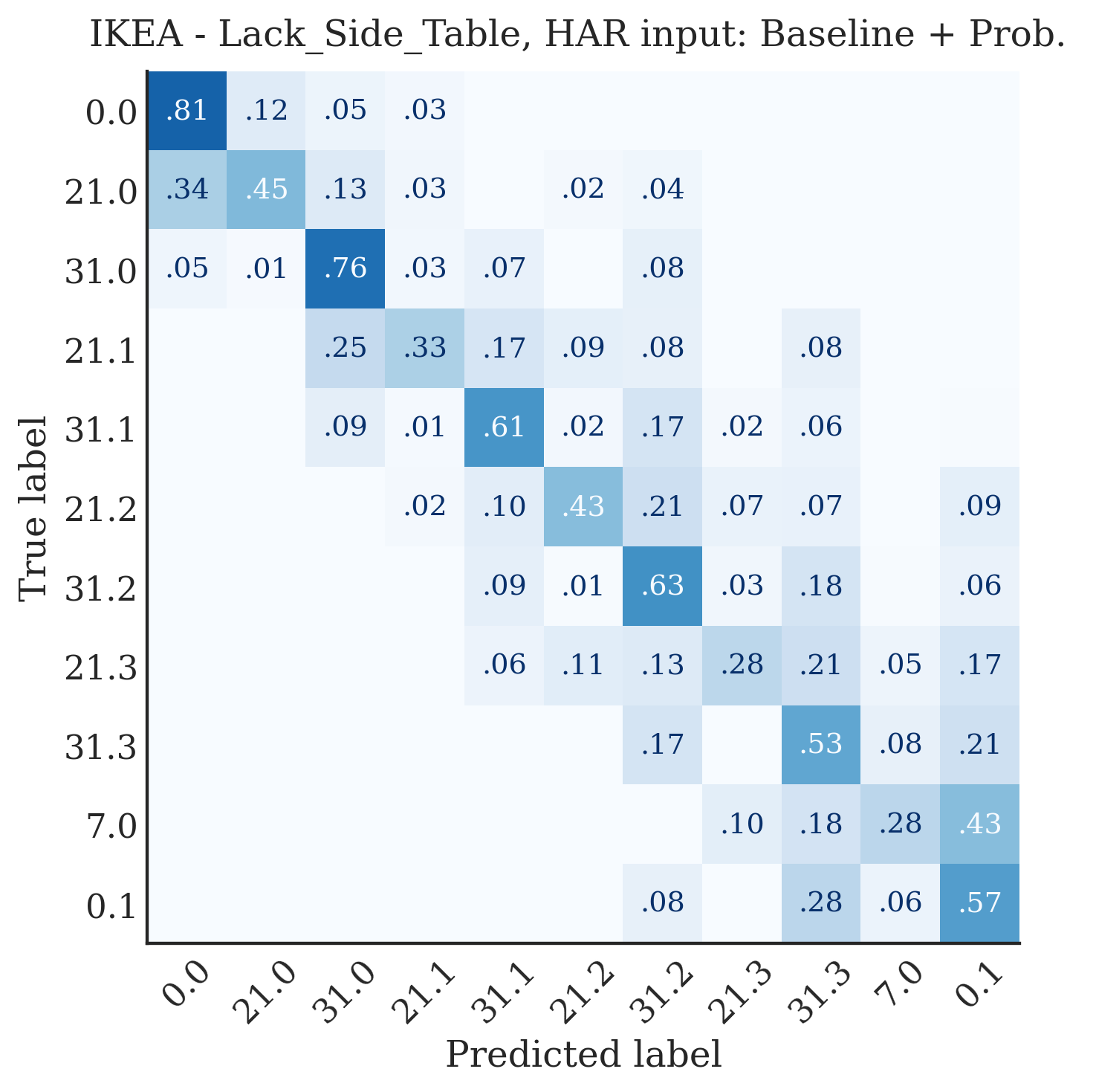}
    \end{subfigure}
    \hfill
    \begin{subfigure}{\confmatsize\linewidth}
        \centering
        \includegraphics[width=\linewidth]{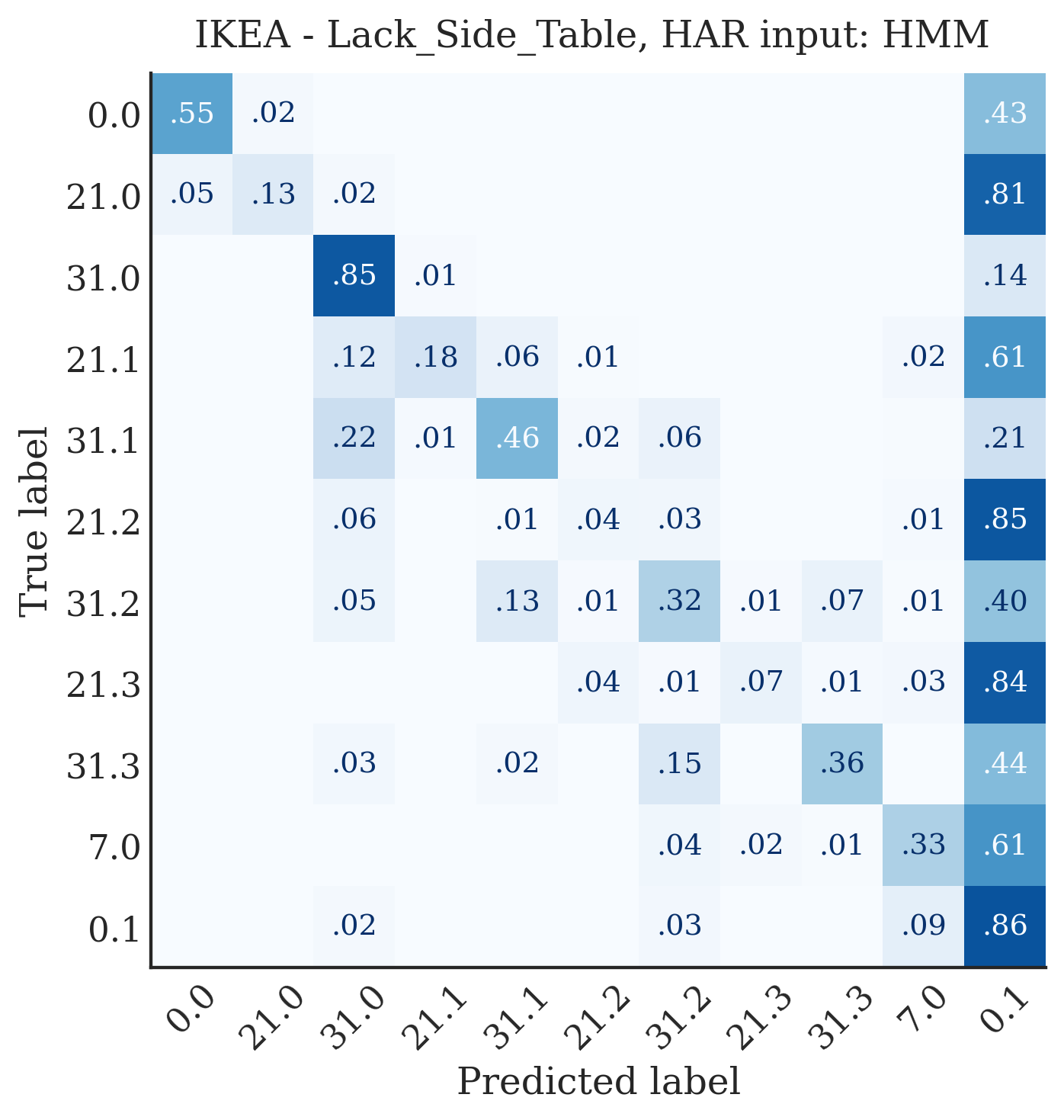}
    \end{subfigure}
    \hfill
    \begin{subfigure}{\confmatsize\linewidth}
        \centering
        \includegraphics[width=\linewidth]{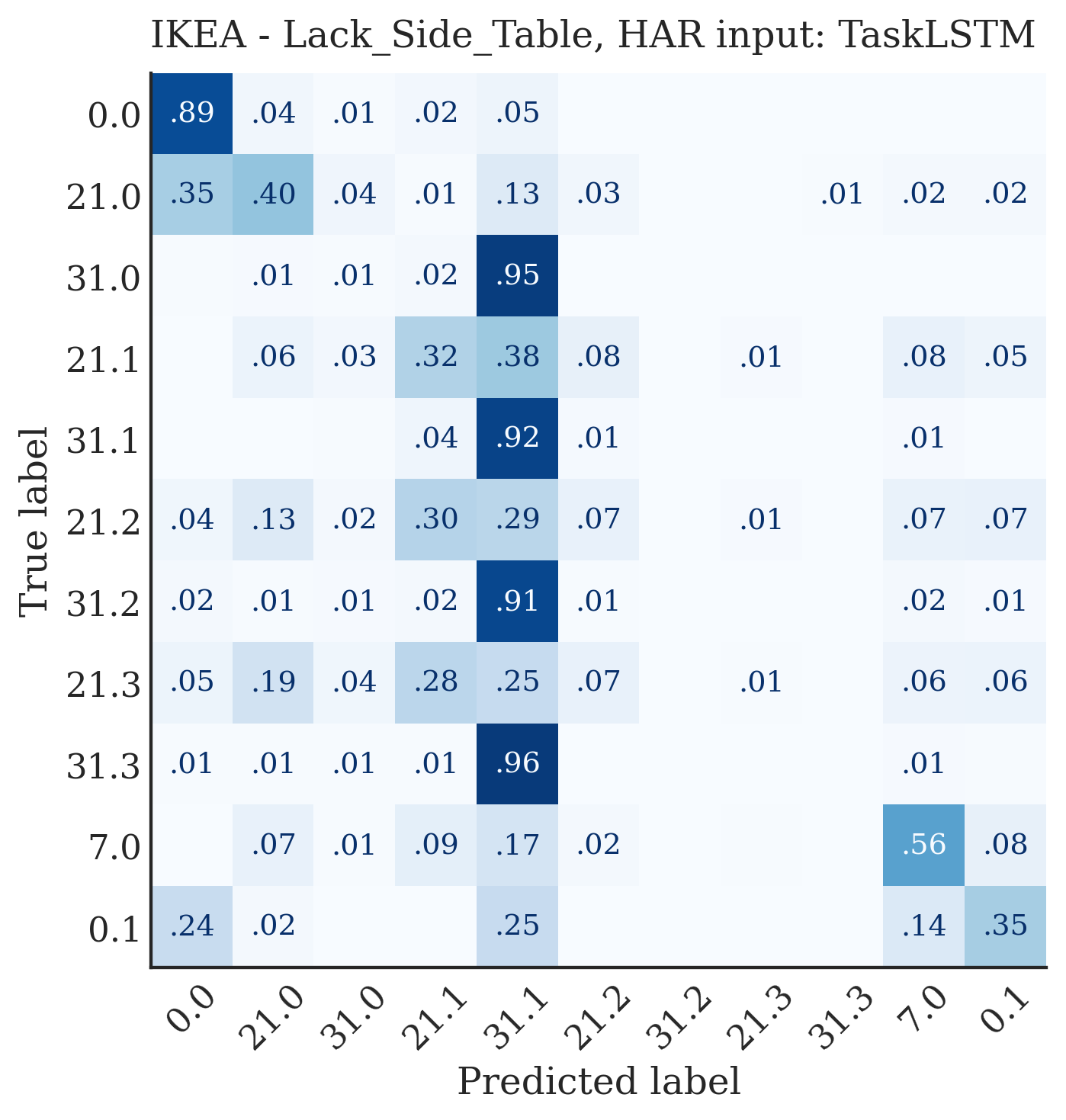}
    \end{subfigure}
    \hfill
    \begin{subfigure}{\confmatsize\linewidth}
        \centering
        \includegraphics[trim={2mm 0mm 2mm 0mm},clip,width=\linewidth]{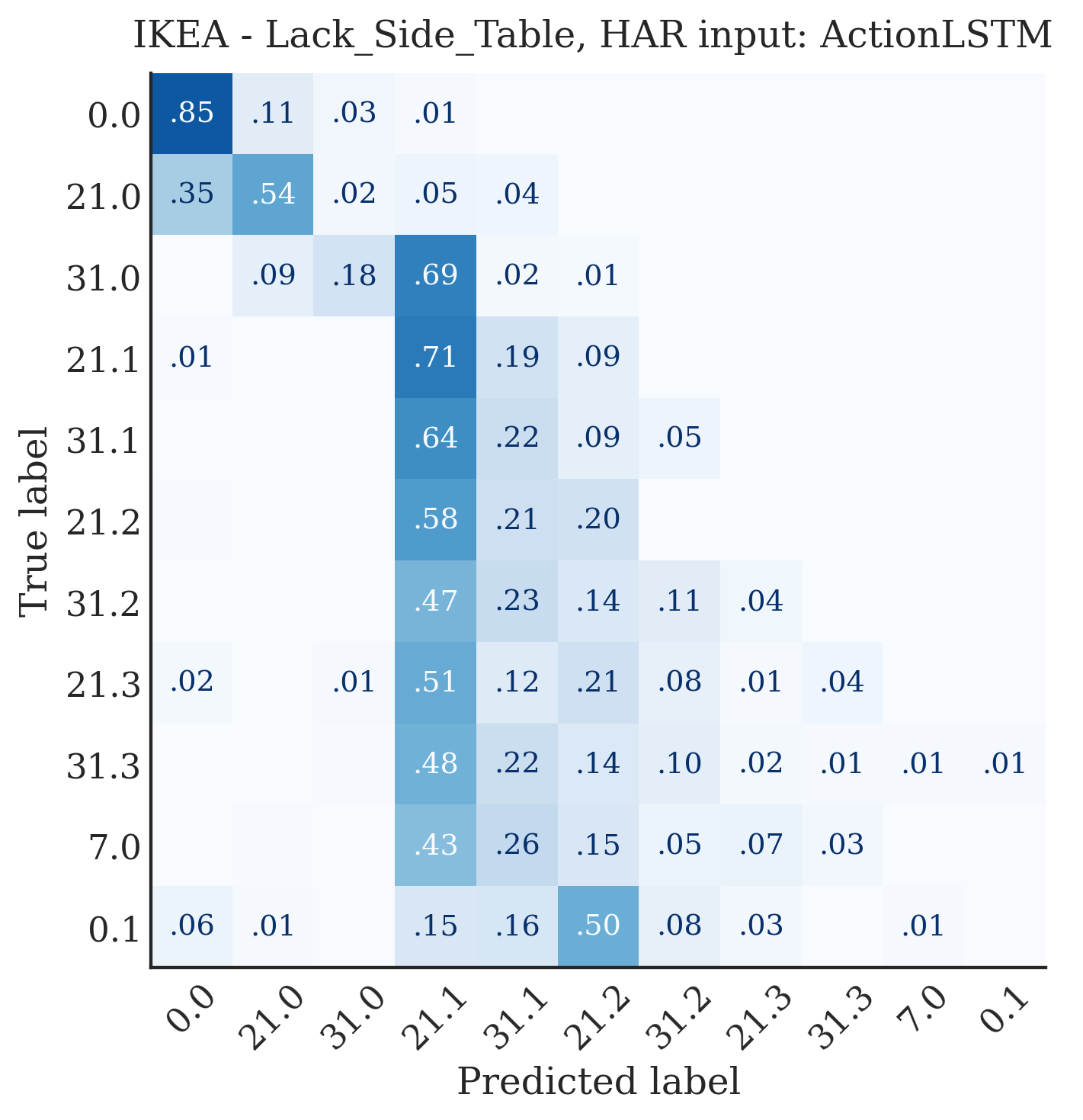}
    \end{subfigure}
    \hfill
    \vspace{-0.005em} 
    \caption{Confusion matrices for IKEA Lack Side Table task state prediction with HAR input.}
    \label{fig:conf_mats_Lack_Side_Table_har}
\end{figure*}

\begin{figure*}[tb]
    \centering
    \includegraphics[width=\linewidth]{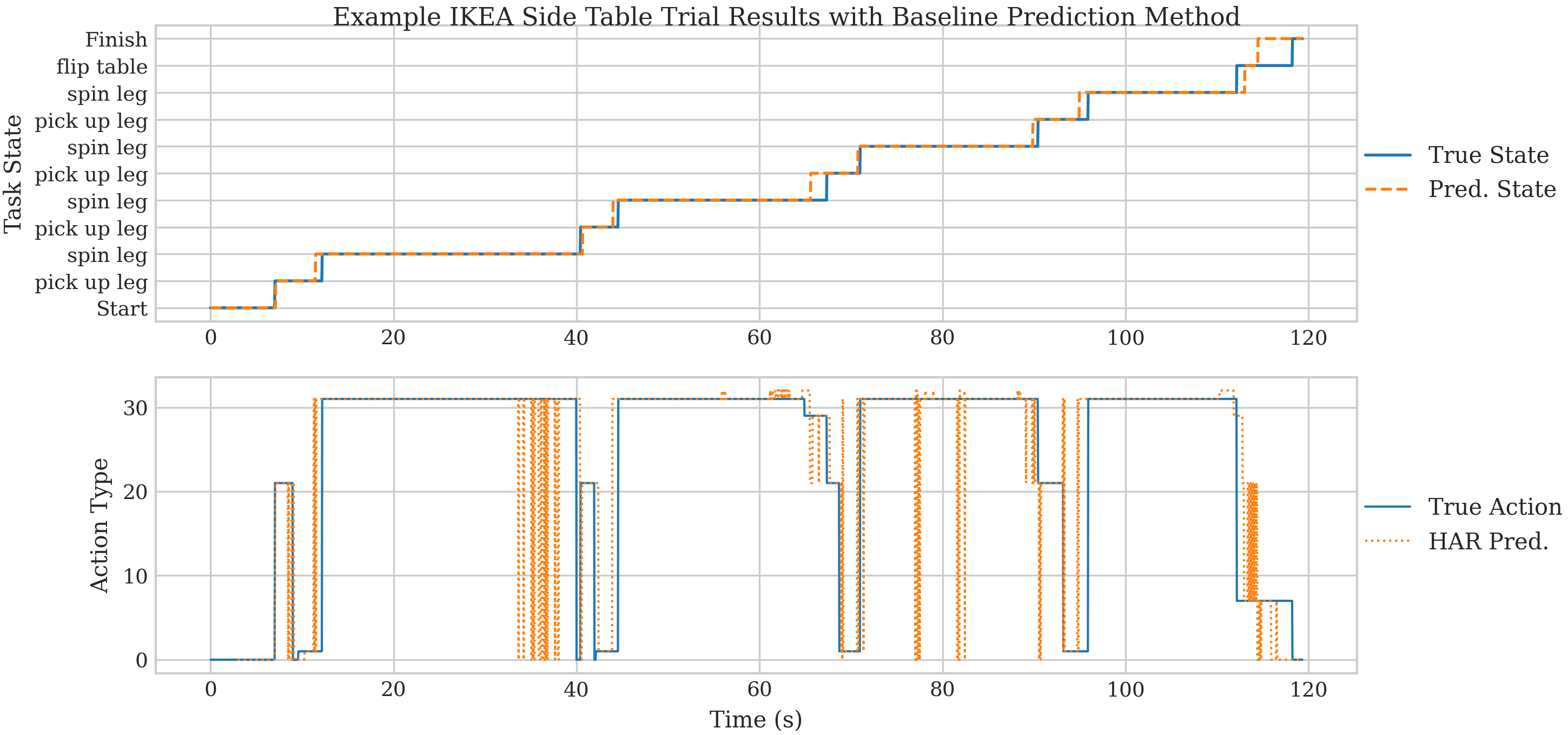}
    \caption{Example state predictions from baseline method and action inputs from the ST-GCN HAR model.}
    \label{fig:example_trial_output}
\end{figure*}

The output from the models is fed as input to the different state prediction methods for each of the test trials. The results of these experiments are shown in Table~\ref{tab:results_har}. It can be seen that the TaskLSTM model again performs best for the HA4M dataset with an F1 of 0.57, while the Baseline+Prob. method is best for the IKEA dataset with an F1 of 0.46. The confusion matrices in Figs.~\ref{fig:conf_mats_ha4m_har}~\&~\ref{fig:conf_mats_Lack_Side_Table_har} show further insight into the models' performances. An example plot of the output from the Baseline state predictor model against the true value, along with action input data is shown in Fig.~\ref{fig:example_trial_output}.



\begin{table}[]
    \centering
    \caption{State recognition metrics with HAR input for HA4M task, IKEA tasks average and average across all tasks.}
    \resizebox{\linewidth}{!}{%
    \begin{tabular}{c|cc|cc|cc}
        \multicolumn{1}{c}{} & \multicolumn{6}{c}{Task Type} \\
        & \multicolumn{2}{c|}{HA4M} 
        & \multicolumn{2}{c|}{IKEA ASM} 
        & \multicolumn{2}{c}{Average} \\
        Method
        & Acc & F1 
        & Acc & F1 
        & Acc & F1 \\ \hline

        Baseline & 42.4 & 41.4 & 40.1 & 36.6 & 40.6 & 37.8 \\
        Baseline + Prob. & 44.9 & 46.3 & \textbf{47.9} & \textbf{46.4} & \textbf{47.2} & \textbf{46.3} \\
        HMM & 51.9 & 50.4 & 38.7 & 39.8 & 42.0 & 42.5 \\
        Task LSTM & \textbf{56.6} & \textbf{57.3} & 29.3 & 21.0 & 36.1 & 30.1 \\
        Action LSTM & 45.5 & 46.0 & 17.8 & 18.2 & 24.7 & 25.1 \\

    \end{tabular}%
    }
    \label{tab:results_har}
\end{table}

\section{Discussion}\label{sec:discussion}
Overall, the results show the challenges in accurately tracking the state of a task. While many of the methods follow the general progression of the task, shown through the broadly diagonal confusion matrices, significant noise around the diagonal highlights the difficulty in accurately predicting the state. Further insights from the confusion matrices can be seen. The vertical lines indicate the method getting stuck at a certain action, while horizontal lines indicate the model skipping ahead. In particular, tasks with repeated actions of the same type are difficult to distinguish (e.g. in the HA4M task), while repeated action sequences are also challenging as in the IKEA tasks.

In general the datasets here present results on more complex situations than those otherwise seen in literature. The datasets contain users assembling products naturally, which means assembly actions and steps are sometimes repeated, skipped or slightly reordered. Furthermore, the tasks are challenging with the same atomic actions being mapped to multiple stages of the assembly. Direct comparison with previous works can be challenging, as metrics reported are often either just for HAR performance or task-based metrics, such as wait time. One of the most comparable works looks at the Assembly 101 dataset and investigates a custom state recognition method as well as an HMM method. Dwivedi et al.'s method achieves 50.9\% accuracy, and the HMM 23.5\%, with both results decreasing similarly as input noise is added to the system~\cite{dwivedi_prediction_2024}. Various works have looked at the similar topic of intent prediction with potentially transferable findings for future work. The inclusion of transition states is important to improve HAR accuracy~\cite{zhang_hybrid_2021}, while task constraints improved HMM intent prediction accuracy from 77.3\% to 90.6\%~\cite{qu_prediction_2025}. When adding multimodal data to HAR inputs, intent prediction accuracy improved from 76\% to 84\%~\cite{xiao_intelligent_2025}, another aspect to explore further.

It is clear inclusion of duration data helps provide robust state tracking, especially where repeated actions occur or to allow recovery if actions are otherwise skipped. Furthermore, accurate HAR has a clear impact on state tracking robustness. Further research is needed on how to make these methods more robust, especially by including additional data such as object state tracking. A key challenge is presenting a robust and generalisable solution which can be deployed with minimal model training or logic configuration.

\section{Conclusion}\label{sec:conclusion}
This paper has presented an investigation into task state modelling based on HAR input data. The work feeds into the goal of fluent collaboration between user and robot in pursuit of Industry 5.0, where the robot should perceive task state based on an understanding of what user actions have been completed. In future work, this will allow the robot partner to adapt its actions to the user's current need. The results highlight the challenges in robust state prediction in complex tasks, where tasks contain repeated actions and users naturally introduce variance to the action sequence. Logic-based approaches performed broadly well, while LSTM and HMM approaches performed better in certain circumstances but were prone to failure when noise and variation were introduced.

\section*{Acknowledgements}
This project has received funding from the European Union's Horizon Europe research and innovation programme under Grant Agreement no. 101059903 and 101135708. 

\bibliographystyle{IEEEtran}
\bibliography{IEEEabrv, reference.bib}

\end{document}